\documentclass{article}

\usepackage[preprint]{corl_2024} 
\usepackage{graphicx}
\usepackage{booktabs}
\usepackage{multirow}
\usepackage{amsmath}
\usepackage{caption}
\usepackage{subcaption}
\usepackage{xcolor} 
\usepackage{colortbl}

\usepackage{adjustbox}
\usepackage{tabularx}
\usepackage{tabularray}
\usepackage{amssymb}
\usepackage{pifont}
\usepackage{enumitem}
\usepackage{hyperref}
\hypersetup{linkcolor=red}

\newcommand{\cmark}{\ding{51}}%

\title{\textsc{DriveVLM}: The Convergence of Autonomous Driving and Large Vision-Language Models}

\author{
Xiaoyu Tian$\phantom{}^{1}\phantom{}^{*}$\hspace{5pt}
Junru Gu$\phantom{}^{1}\phantom{}^{*}$\hspace{5pt}
Bailin Li$\phantom{}^{2}\phantom{}^{*}$\hspace{5pt}
Yicheng Liu$\phantom{}^{1}\phantom{}^{*}$\hspace{5pt}
Yang Wang$\phantom{}^{2}$\hspace{5pt}
Zhiyong Zhao$\phantom{}^{2}$\hspace{5pt}
\vspace{2pt}
\and
Kun Zhan$\phantom{}^{2}$\hspace{5pt}
Peng Jia$\phantom{}^{2}$\hspace{5pt}
Xianpeng Lang$\phantom{}^{2}$\hspace{5pt}
Hang Zhao$\phantom{}^{1}\phantom{}^{\text{†}}$\vspace{5pt} \\
$\phantom{}^1$IIIS, Tsinghua University\hspace{5pt}
$\phantom{}^2$Li Auto\hspace{5pt}
}

\begin{document}
\maketitle

{\begin{NoHyper}\let\thefootnote\relax\footnotetext{$^{*}$Equal contribution. Listing order is random.}\end{NoHyper}}
{\begin{NoHyper}\let\thefootnote\relax\footnotetext{$^{\text{†}}$Corresponding to: hangzhao@mail.tsinghua.edu.cn}\end{NoHyper}}
\begingroup
\renewcommand{\thefootnote}{}
\footnotetext{Project page: \href{https://tsinghua-mars-lab.github.io/DriveVLM/}{https://tsinghua-mars-lab.github.io/DriveVLM/}}
\addtocounter{footnote}{-1}
\endgroup

\begin{abstract}
A primary hurdle of autonomous driving in urban environments is understanding complex and long-tail scenarios, such as challenging road conditions and delicate human behaviors. We introduce DriveVLM, an autonomous driving system leveraging Vision-Language Models (VLMs) for enhanced scene understanding and planning capabilities. DriveVLM integrates a unique combination of reasoning modules for scene description, scene analysis, and hierarchical planning. Furthermore, recognizing the limitations of VLMs in spatial reasoning and heavy computational requirements, we propose DriveVLM-Dual, a hybrid system that synergizes the strengths of DriveVLM with the traditional autonomous driving pipeline. Experiments on both the nuScenes dataset and our SUP-AD dataset demonstrate the efficacy of DriveVLM and DriveVLM-Dual in handling complex and unpredictable driving conditions. Finally, we deploy the DriveVLM-Dual on a production vehicle, verifying it is effective in real-world autonomous driving environments.
\end{abstract}

\section{Introduction}
Autonomous driving, with its great promise to revolutionize transportation, has been an active research area over the past two decades. 
A primary hurdle to a fully autonomous driving system is scene understanding~\cite{barabas2017current}, which involves navigating complex, unpredictable scenarios such as adverse weather, intricate road layouts, and unforeseen human behaviors.

Existing autonomous driving systems, typically comprising 3D perception, motion prediction, and planning, struggle with these scene understanding challenges. Specifically, 3D perception~\cite{qi2017pointnet,lang2019pointpillars,wang2022detr3d,li2022bevformer} is limited to detecting and tracking familiar objects, omitting rare objects and their unique attributes; motion prediction~\cite{vectornet,tnt,liu2021multimodal,gu2021densetnt,nayakanti2023wayformer} and planning~\cite{pomerleau1988alvinn,bansal2018chauffeurnet,li2023uncertainty} focus on trajectory-level actions, often neglecting the decision-level interactions between objects and vehicles.

We introduce \textbf{DriveVLM}, a novel autonomous driving system that aims at these scene understanding challenges, capitalizing on the recent Vision-Language Models (VLMs)~\cite{lynx,xu2023drivegpt4,zhu2023minigpt,liu2023llava} which have demonstrated exceptional prowess in visual comprehension and reasoning.
Specifically, DriveVLM contains a Chain-of-Though (CoT) process with three key modules: \textit{scene description}, \textit{scene analysis}, and \textit{hierarchical planning}. The scene description module linguistically depicts the driving environment and identifies critical objects in the scene; the scene analysis module delves into the characteristics of the critical objects and their influence on the ego vehicle; the hierarchical planning module formulates plans step-by-step, from meta-actions and decision descriptions to waypoints.
These modules respectively correspond to the components of the traditional \textit{perception-prediction-planning} pipeline, but the difference is that these modules tackle \textit{object perception}, \textit{intention-level prediction} and \textit{task-level planning}, which were extremely challenging to cope with in the past.

While VLMs excel in visual understanding, they have limitations in spatial grounding and reasoning, and their computational intensity poses challenges for onboard inference speed. 
Therefore we further propose \textbf{DriveVLM-Dual}, a hybrid system that combines the strengths of both DriveVLM and traditional systems. DriveVLM-Dual optionally integrates DriveVLM with traditional 3D perception and planning modules, such as 3D object detectors, occupancy networks, and motion planners, enabling the system to achieve 3D grounding and high-frequency planning abilities. This dual system design, akin to the human brain's slow and fast thinking processes, adapts efficiently to varying complexity in driving scenarios.

Meanwhile, we formally define the scene understanding and planning (SUP) task, and propose new evaluation metrics to assess the scene analysis and meta-action planning capabilities of DriveVLM and DriveVLM-Dual.
We carry out a comprehensive data mining and annotation pipeline to construct an in-house SUP-AD dataset for the SUP task.
Extensive experiments on both the nuScenes dataset and our own dataset demonstrate the superior performance of DriveVLM, particularly in few-shot scenarios. Furthermore, DriveVLM-Dual exceeds state-of-the-art end-to-end motion planning methods. We have also deployed the model on a production vehicle, confirming that DriveVLM-Dual is effective in real-world autonomous driving environments. Additionally, we have included a demo in the supplementary materials.

\noindent In summary, the contribution of this paper is three-fold:
\vspace{-2mm}
\begin{enumerate}[leftmargin=*,]
    \item We introduce DriveVLM, a novel autonomous driving system that leverages VLMs for effective scene understanding and planning. We further introduce DriveVLM-Dual, a hybrid system that incorporates DriveVLM and a traditional autonomous pipeline, which achieves improved spatial reasoning and real-time planning capabilities. 
    \item We present a comprehensive data mining and annotation pipeline to construct a scene understanding and planning dataset (SUP-AD), together with metrics for evaluation. 
    \item We have successfully deployed DriveVLM-Dual system in a production vehicle and test various effective strategies for accelerating VLM deployment in real driving scenarios.
\end{enumerate}

\vspace{-2pt}
\section{Related Works}
\vspace{-4pt}
\noindent\textbf{Vision-Language Models (VLMs).} Recently, there has been a surge in research on large Vision-Language Models (VLMs), exemplified by works such as MiniGPT-4~\cite{zhu2023minigpt}, LLaVA~\cite{liu2023llava}, Qwen-VL~\cite{Qwen-VL}, and others~\cite{dai2023instructblip, lynx, zhang2023internlmxcomposer, wang2023cogvlm}. 
VLMs can be used in various scenarios, especially robotics~\cite{driess2023palme,rt-1,rt22023arxiv,voxposer,rt-x}, where VLMs output corresponding actions that can be high-level instructions~\cite{driess2023palme} or low-level robot actions~\cite{rt22023arxiv}. DriveVLM focuses on utilizing VLMs to assist in autonomous driving, thereby establishing a novel framework. A Concurrent work~\cite{xu2023drivegpt4} shares a similar motivation.

\noindent\textbf{Learning-based Planning.}
The integration of learning frameworks into motion planning has been an active area of research since Pomerleau~\cite{pomerleau1988alvinn} pioneering contributions. One promising line of work is Reinforcement learning and imitation learning~\cite{chekroun2023gri,chen2021learning,toromanoff2020end}. These methods can learn an end-to-end planning policy that directly maps raw sensory inputs to control actions~\cite{toromanoff2020end}. Several works~\cite{zeng2019end,wei2021perceive,hu2021safe,casas2021mp3} improve interpretability by explicitly building dense cost maps derived from learning-based modules. A recent trend involves training multiple blocks in an end-to-end fashion~\cite{hu2021safe,casas2021mp3,hu2023planning,renz2022plant}. These methods enhance overall performance, but rely on backpropagation from future trajectory predictions loss in a less interpretable decision-making process~\cite{chen2023end}. 


\noindent\textbf{Driving Caption Datasets.}
Recent works~\cite{hu2023gaia,xu2023drivegpt4,yang2023survey} argue that language captions are an important medium to connect human knowledge with the driving objective, helping to inform decisions and actions. Refer-KITTI~\cite{wu2023referring} annotates objects in the KITTI dataset~\cite{geiger2013vision} with language prompts that can reference a collection of objects. Talk2Car~\cite{deruyttere2019talk2car}, NuPrompt~\cite{wu2023language} and nuScenes-QA~\cite{qian2023nuscenes} introduce free-form captions and QA annotation to the nuScenes dataset~\cite{caesar2020nuscenes}. BDD-X~\cite{kim2018textual} and BDD-OIA~\cite{xu2020explainable} offer datasets with language explanations for the ego vehicle’s actions or traffic scenarios~\cite{sachdeva2023rank2tell}~\cite{malla2023drama}. These datasets offer scenes for natural language use, but lack sufficient data on critical safety scenarios in self-driving systems.


\begin{figure*}[t]
    \centering
    \includegraphics[width=0.98\linewidth]{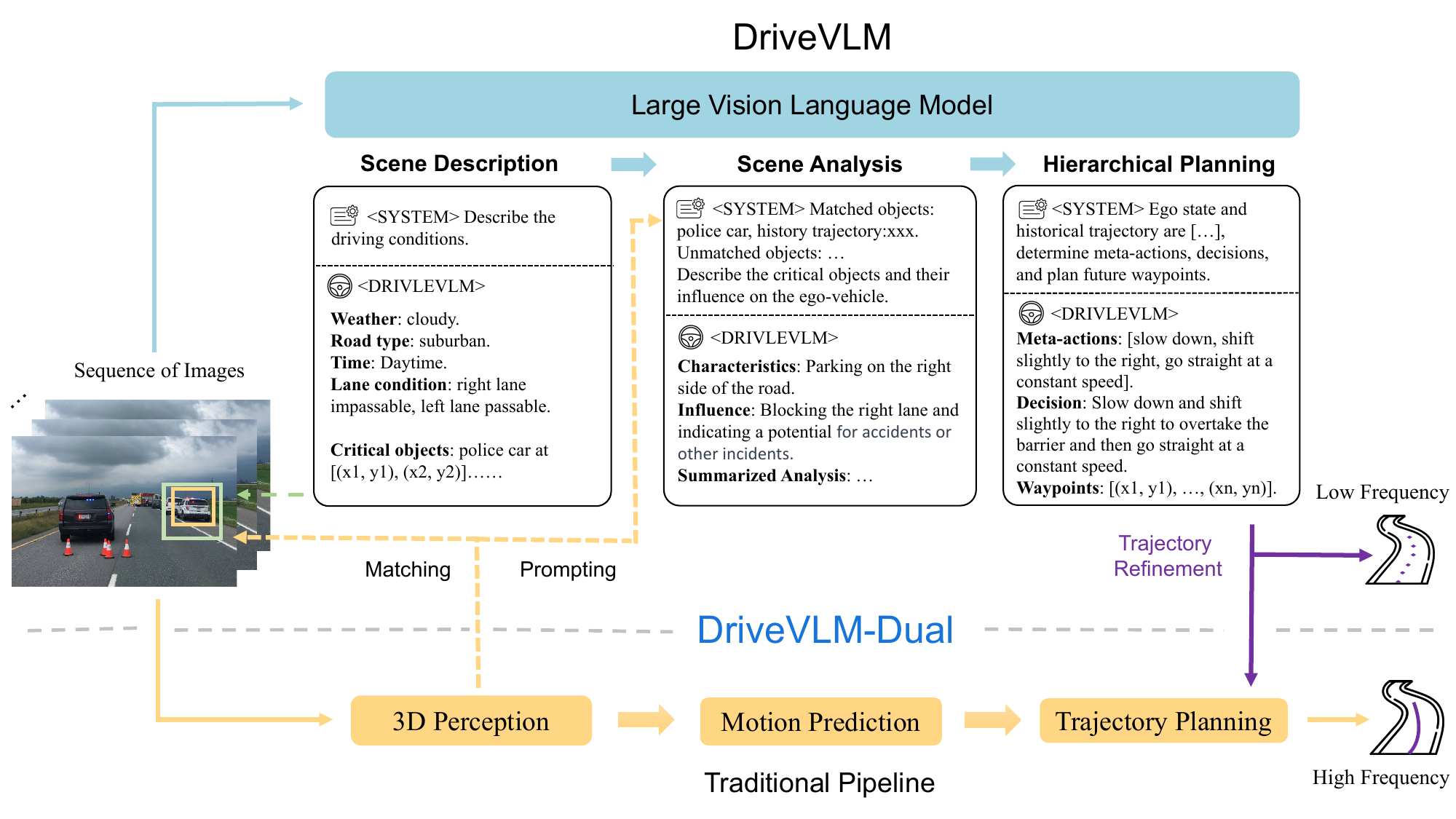}
    \vspace{-1em}
    \caption{\textbf{DriveVLM and DriveVLM-Dual model pipelines.} DriveVLM takes images as input and, through a Chain-of-Thought (CoT) mechanism, outputs scene description, scene analysis, and hierarchical planning results. DriveVLM-Dual further incorporates traditional 3D perception and trajectory planning modules to achieve spatial reasoning capability and real-time trajectory planning.}
    
    \label{fig:pipeline}
    \vspace{-10pt}
\end{figure*}

\section{DriveVLM}
\vspace{-0.5em}
\subsection{Overview}
\vspace{-0.5em}
The overall pipeline of DriveVLM is illustrated in Figure~\ref{fig:pipeline}. A sequence of images is processed by a Vision Language Model (VLM) to perform a special chain-of-thought (CoT)~\cite{wei2022chain} reasoning to derive the driving planning results.
The architecture of DriveVLM involves a vision transformer encoder~\cite{vit} and a Large Language Model (LLM). The vision encoder produces image tokens; then an attention-based extractor aligns these tokens with the LLM.
The reasoning process can be divided into three modules: scene description (Section~\ref{sec:scene-description}), scene analysis (Section~\ref{sec:critical-objects}), and hierarchical planning (Section~\ref{sec:hierarchical-planning}). 


For real-world deployment, we propose a hybrid system, DriveVLM-Dual, in Section~\ref{sec:drivevlm-dual}, which combines DriveVLM and the traditional autonomous driving pipeline, leveraging the strengths of both approaches.


\subsection{Scene Description}
\label{sec:scene-description}
The scene description module identifies driving environment description and critical objects.

\noindent\textbf{Environment Description.}
Driving environments, such as weather and road conditions, have a non-negligible impact on driving difficulty. Therefore, the model is first prompted to output a linguistic description $E$ of the driving environment, including several conditions: $E = \{E_\text{weather}, E_\text{time}, E_\text{road}, E_\text{lane}\},$ each representing a crucial aspect of the driving environment. The weather component, \(E_{\text{weather}}\), spans conditions from sunny to snowy, affecting visibility and traction. The time component, \(E_{\text{time}}\), distinguishes between daytime and nighttime, impacting driving strategies due to visibility changes. Road types, \(E_{\text{road}}\), such as urban or highway, introduce different challenges, while lane conditions, \(E_{\text{lane}}\), focus on current lane positioning and possible maneuvers, crucial for safe driving decisions.


\noindent\textbf{Critical Object Identification.}
In addition to environmental conditions, various objects in driving scenarios significantly influence driving behaviors. 
Unlike traditional autonomous driving perception modules, which detect all objects within a specific range, we solely focus on identifying \textit{critical objects} that are most likely to influence the current scenario, inspired by human cognitive processes during driving. 
Each \textit{critical object}, denoted as $O_c$, contains two attributes: the object category $c$ and its approximate bounding box coordinates $b(x1, y1, x2, y2)$ on the image. 
The category and coordinates are mapped to their corresponding language $token\_id$ in the language modality, enabling seamless integration into the following modules. 
Moreover, taking advantage of the pre-trained vision encoder, DriveVLM can identify long-tail \textit{critical objects} that may elude typical 3D object detectors, such as road debris or unusual animals. 
\vspace{-4pt}

\subsection{Scene Analysis}
\vspace{-2pt}

In the traditional autonomous driving pipeline, the prediction module typically concentrates on forecasting the future trajectories of objects. The emergence of advanced vision-language models has provided us with the ability to perform a more comprehensive analysis of the current scene. 
The scene-level analysis summarizes all the critical objects together with the environmental description. This summary gives a comprehensive understanding of the scene, and is fed into the following planning module.
\label{sec:critical-objects}

\noindent\textbf{Critical Object Analysis.}
DriveVLM characterizes critical objects in three aspects: \textbf{static attributes} $C_s$, \textbf{motion states} $C_m$, and \textbf{particular behaviors} $C_b$. 
Static attributes $C_s$ describe inherent properties of objects, such as a roadside billboard's visual cues or a truck's oversized cargo, which are critical in preempting and navigating potential hazards. 
Motion states $C_m$ describe an object's dynamics over a period, including position, direction, and action—characteristics that are vital in predicting the object's future trajectory and potential interactions with the ego vehicle.
Particular behaviors $C_b$ refer to special actions or gestures of an object that could directly influence the ego vehicle's next driving decisions. 
We do not require the model to analyze all three characteristics for all objects. In practice, only one or two characteristics apply to a critical object. Upon analyzing these characteristics, DriveVLM then predicts the potential influence $I$ of each critical object on the ego vehicle. 


\vspace{-4pt}
\subsection{Hierarchical Planning}
\vspace{-2pt}
\label{sec:hierarchical-planning}
The scene-level summary is then combined with the route, ego pose and velocity to form a prompt for planning.
Finally, DriveVLM progressively generates driving plans, in three stages: meta-actions, decision description, and trajectory waypoints.

\noindent\textbf{Meta-actions $\boldsymbol{A}$.}
A meta-action, denoted as $a_i$, represents a short-term decision of the driving strategy. These actions fall into 17 categories, including but not limited to acceleration, deceleration, turning left, changing lanes, minor positional adjustments, and waiting. To plan the ego vehicle's future maneuver over a certain period, we generate a sequence of meta-actions. 


\noindent\textbf{Decision Description $\boldsymbol{D}$.}
Decision description $D$ articulates the more fine-grained driving strategy the ego vehicle should adopt. It contains three elements: Action $\mathcal{A}$, Subject $\mathcal{S}$, and Duration $\mathcal{D}$. \textit{Action} pertains to meta actions such as `turn', `wait', or `accelerate'. \textit{Subject} refers to the interacting object, such as a pedestrian, a traffic signal, or a specific lane. \textit{Duration} indicates the temporal aspect of the action, specifying how long it should be carried out or when it should start. 



\noindent\textbf{Trajectory Waypoints $\boldsymbol{W}$.} 
Upon establishing the decision description \({D}\), our next phase involves the generation of corresponding trajectory waypoints. These waypoints, denoted by \(W = \{w_1, w_2, ..., w_n\}\), \(w_i=(x_i, y_i)\), depict the vehicle's path over a certain future period with predetermined intervals \(\Delta t\). We map these numerical waypoints into language tokens for auto-regressive generation. 



\vspace{-4pt}
\subsection{DriveVLM-Dual}
\label{sec:drivevlm-dual}
\vspace{-2pt}
To mitigate the challenges of high latency and imprecise spatial and motion understanding in VLMs, we propose DriveVLM-Dual, a collaboration between DriveVLM and the traditional autonomous driving system. This novel approach involves two key strategies: incorporating 3D perception for critical object analysis, and high-frequency trajectory refinement.

\noindent\textbf{Integrating 3D Perception.}
We represent objects detected by a 3D detector as \( O_\text{3D} = \{ c_\text{3D}^i, b_\text{3D}^i \} \), where \( b_\text{3D}^i \) denotes the $i$-th bounding box and $c_\text{3D}^i$ denotes its category. These 3D bounding boxes are then back-projected onto 2D images to derive corresponding 2D bounding boxes \( b_\text{2D}^i \). We conduct IoU matching between these 2D bounding boxes \( b_\text{2D}^i \) and \( b_{c}^j \).  \( b_{c}^j \) are the bounding boxes of previously identified critical objects \( O_\text{critical} = \{ c_{c}^j, b_{c}^j \} \). We classify critical objects that meet a certain approximate IoU threshold and belong to the same category as matched critical objects \( O_{c}^\text{matched} \), defined as 
\begin{align}
O_{c}^\text{matched} = \{ c_{c}^j, b_{c}^j \}, \quad & \text{if } c_{c}^j = c_\text{2D}^i \nonumber \text{ and } \text{aIoU}(b_{c}^j, b_\text{2D}^i) > \tau, \text{ where aIoU}(b_c^j, b_\text{2D}^i) = \frac{S_{b_c^j \cap b_\text{2D}^i}}{S_{b_\text{2D}^i}},
\end{align}
Those critical objects without a corresponding match in the 3D data are noted as \( O_{c}^\text{unmatched} \).

In the scene analysis module, for \( O_{c}^\text{matched} \), the center coordinates, orientations, and historical trajectories of the corresponding 3D objects are used as language prompts for the model, assisting in object analysis. Conversely, for \( O_{c}^\text{unmatched} \), analysis relies solely on the language tokens derived from the image. This design enables DriveVLM-Dual to understand the locations and motions of critical objects more accurately, enhancing the overall performance.

\noindent\textbf{High-frequency Trajectory Refinement.}
To achieve real-time, high-frequency inference capabilities, we integrate it with a conventional planner to form a slow-fast dual system, combining the advanced capabilities of DriveVLM with the efficiency of traditional planning methods.
After obtaining a trajectory from DriveVLM at low frequency, denoted as $W_\text{slow}$, we take it as a reference trajectory for a classical planner for high-frequency trajectory refinement. 
In the case of an optimization-based planner, $W_\text{slow}$ serves as the initial solution for the optimization solver. 
For a neural network-based planner, $W_\text{slow}$ is used as an input query, combined with additional input features $f$, and then decoded into a new planning trajectory denoted as $W_\text{fast}$. 
The formulation of this process can be described as:
\begin{align}
W_{\text{fast}} = \text{Planner}([W_\text{slow}, f]).
\end{align}
This refinement step ensures that the trajectory produced by DriveVLM-Dual (1) achieves higher trajectory quality, and (2) meets real-time requirements. 
In practice, the two branches operate asynchronously in a slow-fast manner, where the planner module in the traditional autonomous driving branch can selectively receive trajectory from the VLM branch as additional input.


\section{Task and Dataset}
\label{sec:dataset} 
To fully exploit the potential of DriveVLM and DriveVLM-Dual in handling complex and long-tail driving scenarios, we formally define a task called \textit{Scene Understanding for Planning} (Section~\ref{sec:task-def}), together with a set of evaluation metrics (Section~\ref{sec:eval-metrics}). Furthermore, we propose a data mining and annotation protocol to curate a scene understanding and planning dataset (Section~\ref{sec:data-mining}).

\begin{figure*}[!ht]
    \centering
    \includegraphics[width=1.0\linewidth]{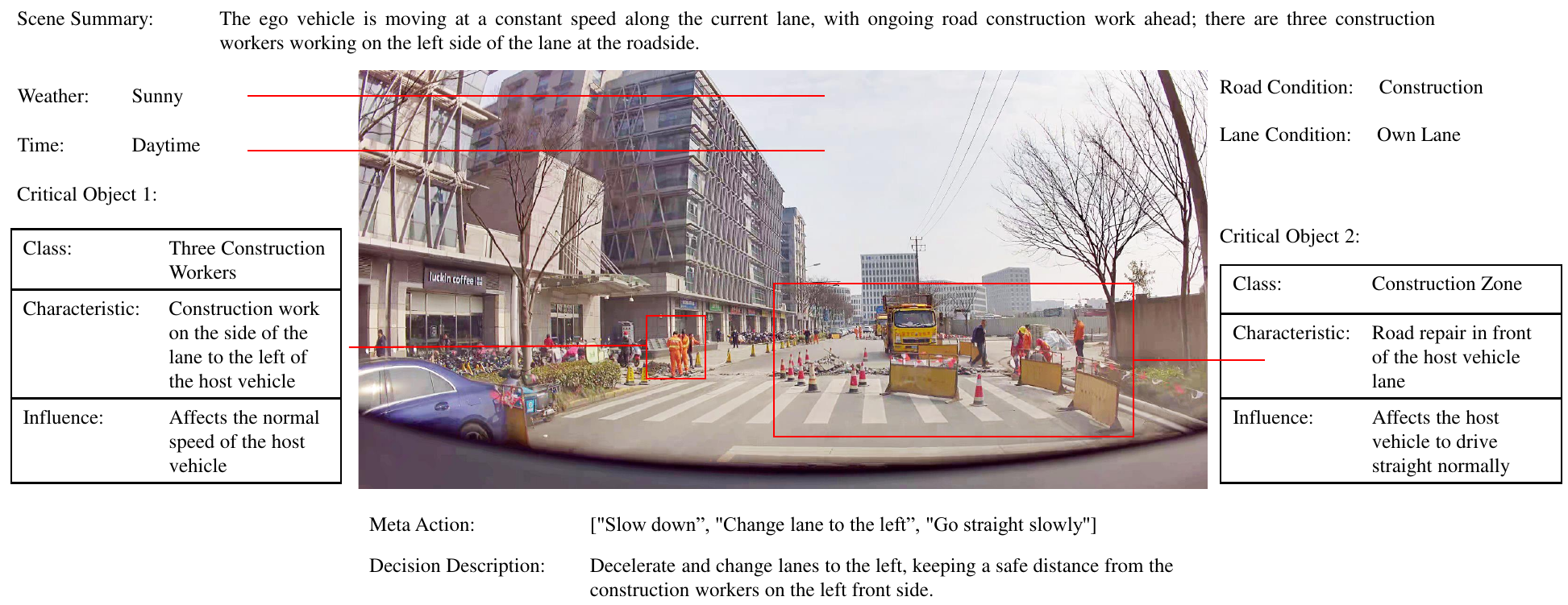}
    \caption{\textbf{An annotated sample of the SUP-AD dataset.}}
    \label{fig:anno-sample}
\vspace{-1em}
\end{figure*}

\subsection{Task Definition}
\label{sec:task-def}
The Scene Understanding for Planning task is defined as follows. The input comprises multi-view videos $\mathcal{V}$ from surrounding cameras and optionally 3D perception results $\mathcal{P}$ from a perception module. The output includes the following components:
\begin{enumerate}
    \item \textbf{Scene Description $E$}: Composed of weather condition $E_\text{weather}$, time $E_\text{time}$, road condition $E_\text{road}$, and lane conditions $E_\text{lane}$.
    \item \textbf{Scene Analysis $S$}: Including object-level analysis  and scene-level summary $S$.
    \item \textbf{Meta Actions $A$}: A sequence of actions representing task-level maneuvers.
    \item \textbf{Decision Description $D$}: A detailed account of the driving decisions.
    \item \textbf{Trajectory Waypoints $W$}: The waypoints outlining the planned trajectory of the ego vehicle.
\end{enumerate}

\vspace{-1em}
\subsection{Evaluation Metrics}
\vspace{-0.4em}
\label{sec:eval-metrics}
To comprehensively evaluate a model's performance, we care about its interpretation of the driving scene and the decisions made. Therefore, our evaluation has two aspects: scene description/analysis evaluation and meta-action evaluation.

\noindent\textbf{Scene Description/Analysis Evaluation.}
Given the subjective nature of human evaluation in scene description, we adopt a structured approach using a pre-trained LLM. This method entails comparing the generated scene description with a human-annotated ground truth description. The ground truth description encompasses structured data such as environmental conditions, navigation, lane information, and critical events with specific objects, verbs, and their influences. The LLM assesses and scores the generated descriptions based on their consistency with the ground truth.

\noindent\textbf{Meta-action Evaluation.}
Meta-actions are a predefined set of decision-making options. A driving decision is formulated as a sequence of meta-actions. Our evaluation method employs a dynamic programming algorithm to compare the model-generated sequences with a manually annotated ground truth sequence. 
The evaluation should also weigh the relative importance of various meta-actions, designating some as `conservative actions' with a lower impact on the sequence's overall context. To increase robustness, we first use the LLM to generate semantically equivalent alternatives to the ground truth sequence to enhance robustness. The sequence with the highest similarity to these alternatives calculates the final driving decision score. 
More details of the proposed metric are available in the Appendix~\ref{sec:appendix-eval}.

\subsection{Dataset Construction}

\label{sec:data-mining}
\begin{figure*}[t]
    \centering
    \includegraphics[width=0.99\linewidth]{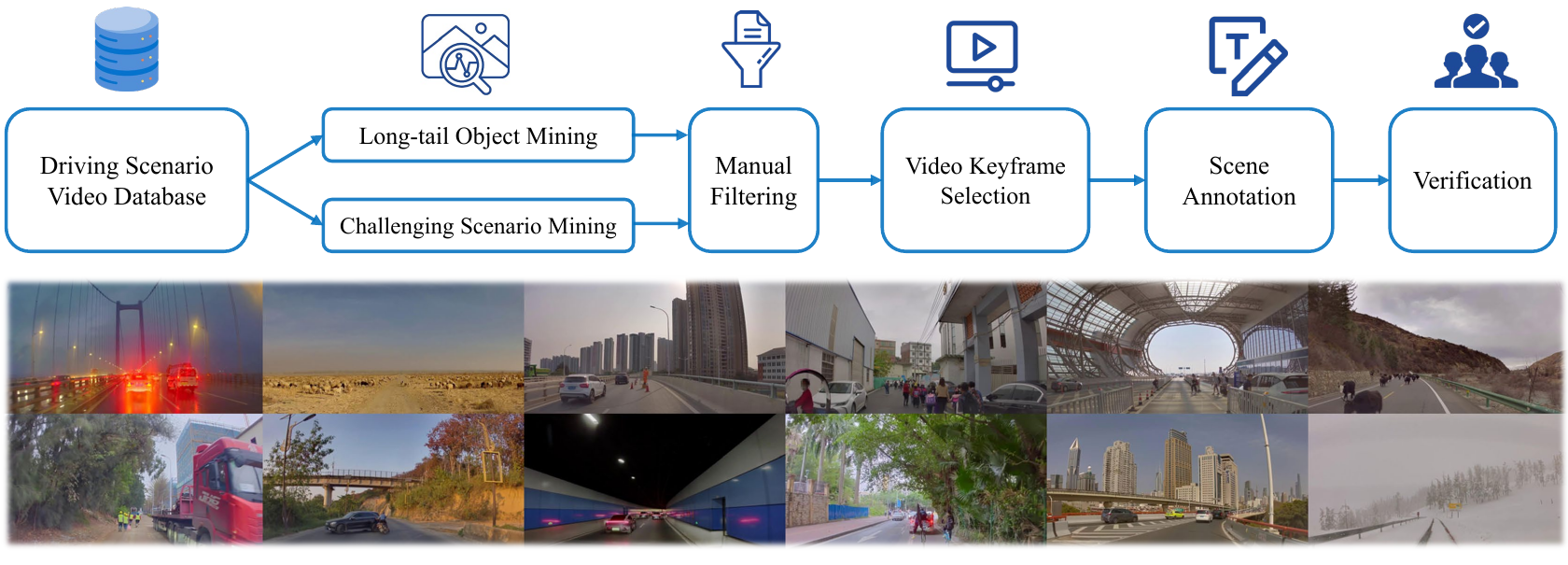}
    \caption{\textbf{The proposed data mining and annotation pipeline for constructing a scene understanding and planning dataset.} Scenario examples randomly sampled from the dataset (below) demonstrate the diversity and complexity of the dataset.}
    \label{fig:dataset_pipeline}
\vspace{-1em}
\end{figure*}

We propose a comprehensive data mining and annotation pipeline, shown in Figure~\ref{fig:dataset_pipeline}, to construct a \textit{Scene Understanding for Planning (SUP-AD) Dataset} for the proposed task. Specifically, we perform long-tail object mining and challenging scenario mining from a large database to collect samples, then we select a keyframe from each sample and further perform scene annotation.
Dataset statistics are available in the Appendix~\ref{sec:appendix-dataset}.

\noindent\textbf{Long-tail Object Mining.}
According to real-world road object distribution, we first define a list of long-tail object categories, such as weird-shaped vehicles, road debris, and animals crossing the road. Next, we mine these long-tail scenarios using a CLIP-based search engine, capable of mining driving data using language queries from a large collection of logs. Following that, we perform a manual inspection to filter out scenes inconsistent with the specified categories.

\noindent\textbf{Challenging Scenario Mining.}
In addition to long-tail objects, we are also interested in challenging driving scenarios, where the driving strategy of the ego vehicle needs to be adapted according to the changing driving conditions. These scenarios are mined according to the variance of the recorded driving maneuvers.

\noindent\textbf{Keyframe Selection.}
Each scene is a video clip, it is essential to identify the `keyframe' to annotate. In most challenging scenarios, a keyframe is the moment before a significant change in speed or direction is required. We select this keyframe 0.5s to 1s earlier than the actual maneuver, based on comprehensive testing, to guarantee an optimal reaction time for decision-making. For scenes that do not involve changes in driving behavior, we select a frame that is relevant to the current driving scenario as the keyframe.

\noindent\textbf{Scene Annotation.}
We employ a group of annotators to perform the scene annotation, including scene description, scene analysis, and planning, except for waypoints, which can be auto-labeled from the vehicle's IMU recordings.
To facilitate scene annotation, we make a video annotation tool with the following features: (1) the annotators can slide the progress bar back and forth to replay any part of a video; (2) while annotating a keyframe, the annotator can draw bounding boxes on the image together with language descriptions; (3) annotators can select from a list of action and decision candidates while annotating driving plans.
Each annotation is meticulously verified by 3 annotators for accuracy and consistency, ensuring a reliable dataset for model training. Figure~\ref{fig:anno-sample} illustrates a sample scenario with detailed annotations.

\section{Experiments}
\vspace{-0.5em}
\subsection{Settings}
We test DriveVLM and DriveVLM-Dual on our proposed SUP-AD dataset and nuScenes dataset~\cite{caesar2020nuscenes}. 

\noindent\textbf{SUP-AD Dataset.} The SUP-AD dataset is a dataset built by our proposed data mining and annotation pipeline. It is divided into train, validation, and test splits with a ratio of $7.5:1:1.5$. We train models on the training split and use our proposed scene description and meta-action metrics to evaluate model performance on the test split. We also employ co-tuning with additional datasets to ensure the generalization of the LLM is not compromised. 

\noindent\textbf{nuScenes Dataset.} The nuScenes dataset is a large-scale driving dataset of urban scenarios with 1000 scenes, where each scene lasts about 20 seconds. 
Following previous works~\cite{hu2023planning,jiang2023vad}, we adopt Displacement Error (DE) and Collision Rate (CR) as metrics to evaluate models' performance.


\begin{table}[t]
\caption{\textbf{Results on the test set of our proposed SUP-AD dataset.} $^\dagger$: Using the official API of GPT-4V. For Lynx and CogVLM, we utilize the training split for fine-tuning purposes. In contrast, for GPT-4V, we employ in-context learning.}
\begin{center}
\resizebox{0.6\columnwidth}{!}{
\begin{tblr}{
  colspec = {l|cc},
  row{1-Z} = {rowsep=0.6pt}, 
  hline{1,2,6} = {0.8pt},
}
Method & Scene Description & Meta-actions \\
    Fine-tuning w/ Lynx~\cite{lynx}              & $0.46$   & $0.15$ \\
    Fine-tuning w/ CogVLM~\cite{wang2023cogvlm}  & $0.49$   & $0.22$   \\
    GPT-4V$^\dagger$~\cite{openai2023gpt4}       & $0.38 $  & $0.19$ \\
    DriveVLM w/ Qwen                              & \textbf{0.71}   & \textbf{0.37}  \\
\end{tblr}
}
\end{center}
\vspace{-1.5em}
\label{tab:main-own}
\end{table}

\begin{table}[t]
\centering
\caption{\textbf{Planning results on the nuScenes validation dataset.} DriveVLM-Dual achieves the best performance. $^\dagger$ denotes cooperating with VAD~\cite{jiang2023vad}.}
\begin{adjustbox}{width=0.7\columnwidth,center}
\begin{tblr}{
  colspec = {l|cccc|cccc},
  rows = {font=\scriptsize},
  row{1-Z} = {rowsep=-2pt},
  hline{1,12} = {0.6pt},
  hline{3,10} = {0.5pt},
  cell{1}{1} = {r=2}{l}, 
  cell{1}{2} = {c=4}{},  
  cell{1}{6} = {c=4}{},  
}
    Method & L2 (m) $\downarrow$ &&&& Collision (\%) $\downarrow$ &&& \\
           & 1s & 2s & 3s & Avg. & 1s & 2s & 3s & Avg. \\
    NMP~\cite{zeng2019end} & - & - & 2.31 & - & - & - & 1.92 & - \\
    SA-NMP~\cite{zeng2019end} & - & - & 2.05 & - & - & - & 1.59 & - \\
    FF~\cite{hu2021safe} & 0.55 & 1.20 & 2.54 & 1.43 & 0.06 & 0.17 & 1.07 & 0.43 \\
    EO~\cite{khurana2022differentiable} & 0.67 & 1.36 & 2.78 & 1.60 & 0.04 & 0.09 & 0.88 & 0.33 \\
    ST-P3~\cite{hu2022st} & 1.33 & 2.11 & 2.90 & 2.11 & 0.23 & 0.62 & 1.27 & 0.71 \\
    UniAD~\cite{hu2023planning} & 0.48 & 0.96 & 1.65 & 1.03 & 0.05 & 0.17 & 0.71 & 0.31 \\
    VAD-Base~\cite{jiang2023vad} & 0.17 & 0.34 & 0.60 & 0.37 & 0.07 & 0.10 & 0.24 & 0.14 \\
    DriveVLM & 0.18 & 0.34 & 0.68 & 0.40 & 0.10 & 0.22 & 0.45 & 0.27 \\
    \textbf{DriveVLM-Dual}$^\dagger$ & \textbf{0.15} & \textbf{0.29} & \textbf{0.48} & \textbf{0.31} & \textbf{0.05} & \textbf{0.08} & \textbf{0.17} & \textbf{0.10} \\
\end{tblr}
\end{adjustbox}
\label{tab:main-nus}
\vspace{-0.5em}
\end{table}

\paragraph{Co-tuning}
To preserve the LLM's generalization capabilities during the fine-tuning process, we employed co-tuning with several additional datasets.
These include the Talk2Car~\cite{deruyttere2019talk2car}, BDDX~\cite{kim2018textual}, Drama~\cite{malla2023drama}, SUTD~\cite{sutd}, and LLAVA~\cite{liu2023llava} datasets. For each dataset, we conducted random sampling in a 1:1 ratio corresponding to the data volume of the SUP-AD and nuScenes datasets. Following this co-tuning approach, we found that the scores on the SUP-AD dataset, under the evaluation metrics of Scene Description and Meta Action, remained virtually unchanged, simultaneously ensuring the preservation of the LLM's original capabilities and its generalization capacity.
\vspace{-0.5em}
\paragraph{Base Model.} We use Qwen-VL~\cite{Qwen-VL} as our default large vision-language model, which exhibits remarkable performance in tasks like question answering, visual localization, and text recognition. It contains a total of 9.6 billion parameters, including a visual encoder (1.9 billion), a vision-language adapter (0.08 billion), and a large language model (Qwen, 7.7 billion). Images are resized to a resolution of $448\times448$ before being encoded by the vision encoder. During training, we randomly select a sequence of images at the current time $T$s, $T-1$s, $T-2$s, and $T-3$s as input. The selected images ensure the inclusion of the current time frame and follow an ascending chronological order.

\vspace{-0.5em}
\subsection{Main Results}
\noindent\textbf{SUP-AD.} We present the performance of our proposed DriveVLM with several large vision-language models and compare them with GPT-4V, as shown in Table~\ref{tab:main-own}. DriveVLM, utilizing Qwen-VL as its backbone, achieves the best performance due to its strong capabilities in question answering and flexible interaction compared to the other open-source VLMs. 
Although GPT-4V exhibits robust capabilities in vision and language processing, its inability to undergo fine-tuning, restricting it solely to in-context learning, often results in the generation of extraneous information during scene description tasks. Under our evaluation metric, the additional information is frequently classified as hallucination, consequently leading to lower scores.


\noindent\textbf{nuScenes.} As shown in Table~\ref{tab:main-nus}, DriveVLM-Dual achieves state-of-the-art performance on the nuScenes planning task when cooperating with VAD. It demonstrates that our method, although tailored for understanding complex scenes, also excels in ordinary scenarios. 

\subsection{Ablation Study}


\noindent\textbf{Model Design.}
To better understand the significance of our designed modules in DriveVLM, we conduct ablations on different combinations of modules, as shown in Table~\ref{tab:abl-nus}. The inclusion of critical object analysis enables our model to identify and prioritize important elements in the driving environment, enhancing the decision-making accuracy for safer navigation. Integrating 3D perception data, our model gains a refined understanding of the surroundings and achieves precise predictions.

\noindent\textbf{Traditional AD Pipeline.} To demonstrate the generalization of our dual system design, we test DriveVLM-Dual with different traditional autonomous driving pipelines on the validation set of nuScenes. As illustrated in Table~\ref{tab:diff-dual}, our proposed DriveVLM-Dual adapts well to different traditional AD pipelines. While a standalone MLP method shows a notable performance gap compared to VAD, both variants of DriveVLM-Dual achieve nearly identical performance, underscoring the efficacy and robustness of our dual system design.




\begin{table}[t]
\centering
\caption{\textbf{Ablations of design choices on the validation set of nuScenes.} ``Base'' refers to only indicating the hierarchical planning results without our proposed CoT inference. ``CO'' represents the addition of critical object analysis. ``3D'' denotes the inclusion of 3D perception results as an auxiliary language prompt.}
\vspace{0.5em}
\begin{adjustbox}{width=0.75\columnwidth,center}
\begin{tblr}{
  colspec = {l|ccc|cccc|cccc},
  rows = {font=\scriptsize},
  row{1-Z} = {rowsep=-2pt},
  hline{1,3,6} = {0.6pt},
  cell{1}{1} = {r=2}{c},
  cell{1}{2} = {r=2}{c},
  cell{1}{3} = {r=2}{c}, 
  cell{1}{4} = {r=2}{c}, 
  cell{1}{5} = {c=4}{},  
  cell{1}{9} = {c=4}{},  
}
    ID & Base & CO & 3D & L2 (m) $\downarrow$ &&&& Collision (\%) $\downarrow$ &&& \\
       &      &    &    & 1s & 2s & 3s & Avg. & 1s & 2s & 3s & Avg. \\
    1 & \cmark & & & 0.19 & 0.41 & 0.89 & 0.49 & 0.16 & 0.28 & 0.63 & 0.36 \\
    2 & \cmark & \cmark & & 0.20 & 0.38 & 0.75 & 0.44 & 0.15 & 0.29 & 0.61 & 0.35 \\
    3 & \cmark & \cmark & \cmark & 0.18 & 0.34 & 0.68 & 0.40 & 0.10 & 0.22 & 0.45 & 0.27 \\
\end{tblr}
\end{adjustbox}
\label{tab:abl-nus}
\vspace{-0.5em}
\end{table}

\begin{table}[t]
\centering
\caption{\textbf{Ablations of traditional autonomous driving pipeline in DriveVLM-Dual.} MLP stands for methods similar to AD-MLP~\cite{ad-mlp}.}
\vspace{0.5em}
\begin{adjustbox}{width=0.75\columnwidth,center}
\begin{tblr}{
  colspec = {l|cccc|cccc},
  rows = {font=\scriptsize},
  row{1-Z} = {rowsep=-2pt},
  hline{1,3,9} = {0.6pt},
  hline{5,7} = {0.5pt},
  cell{1}{1} = {r=2}{l},
  cell{1}{2} = {c=4}{}, 
  cell{1}{6} = {c=4}{},  
}
    Method & L2 (m) $\downarrow$ &&&& Collision (\%) $\downarrow$ &&& \\
           & 1s & 2s & 3s & Avg. & 1s & 2s & 3s & Avg. \\
    UniAD~\cite{hu2023planning} & 0.48 & 0.96 & 1.65 & 1.03 & 0.05 & 0.17 & 0.71 & 0.31 \\
    DriveVLM-Dual (UniAD) & 0.17 & 0.37 & 0.63 & 0.39 & 0.08 & 0.18 & 0.35 & 0.20 \\
    MLP & 0.25 & 0.46 & 0.62 & 0.44 & 0.14 & 0.18 & 0.28 & 0.20 \\
    DriveVLM-Dual (MLP) & 0.14 & 0.35 & 0.30 & 0.31 & 0.09 & 0.13 & 0.18 & 0.13 \\
    VAD~\cite{jiang2023vad} & 0.17 & 0.34 & 0.60 & 0.37 & 0.07 & 0.10 & 0.24 & 0.14 \\
    DriveVLM-Dual(VAD) & 0.15 & 0.29 & 0.48 & 0.31 & 0.05 & 0.08 & 0.17 & 0.10 \\
\end{tblr}
\end{adjustbox}
\label{tab:diff-dual}
\vspace{-0.5em}
\end{table}


\subsection{Qualitative Results}

\begin{figure}[!h]
  \centering
  \subfloat[]{\includegraphics[width=0.49\textwidth]{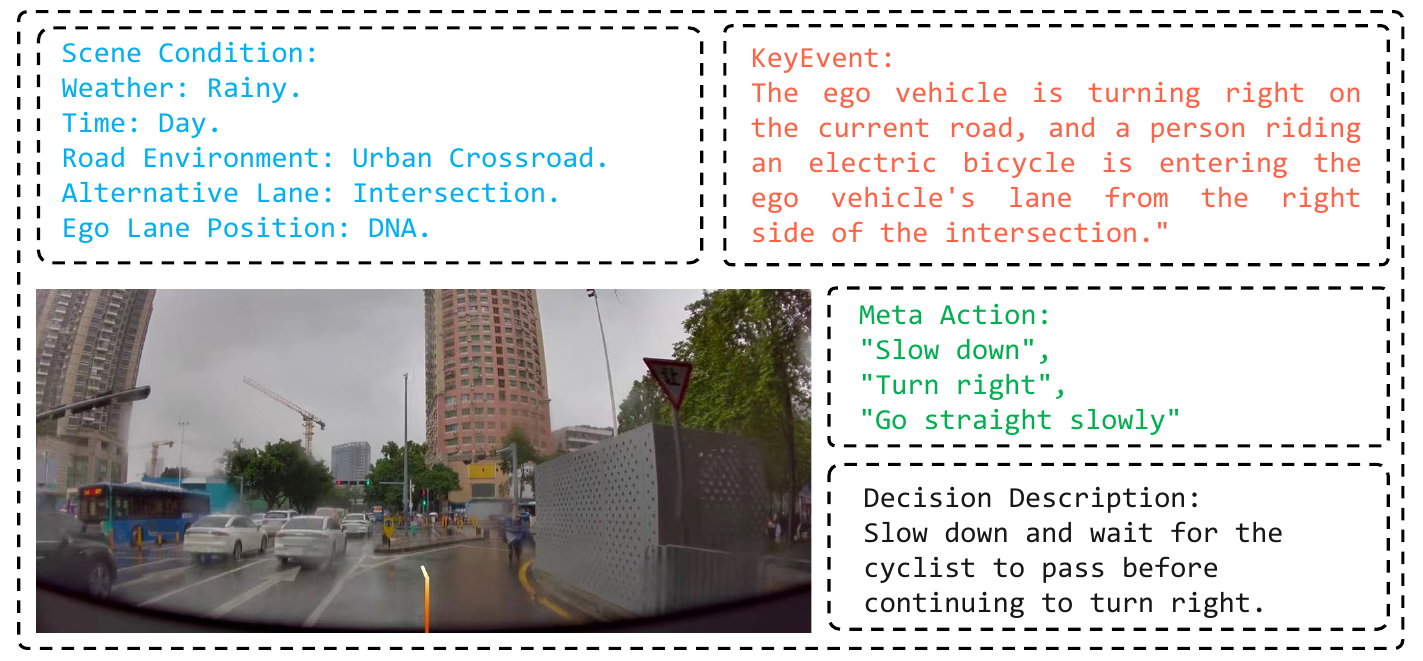}\label{fig:vis1}}
  \hspace{1pt}
  \subfloat[]{\includegraphics[width=0.49\textwidth]{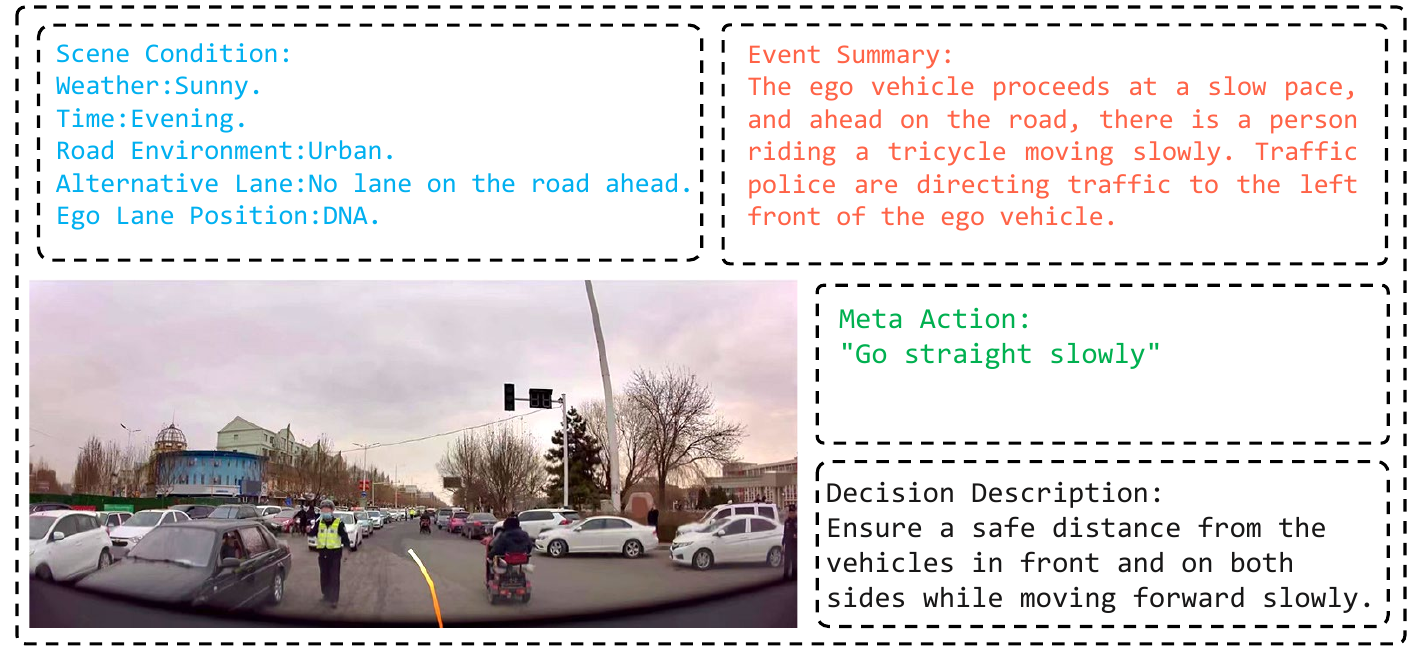}\label{fig:vis2}}
  \vspace{-2pt}
  \caption{\textbf{Qualitative results of DriveVLM.} The orange curves represent the model's planned future trajectories for the next 3 seconds.}

  \label{fig:vis}
\end{figure}

Qualitative results of DriveVLM are shown in Figure~\ref{fig:vis}. In Figure~\ref{fig:vis1}, DriveVLM accurately predicts the current scene conditions and incorporates well-considered planning decisions regarding the cyclist approaching us. 
In Figure~\ref{fig:vis2}, DriveVLM effectively comprehends the gesture of the traffic police ahead, signaling the ego vehicle to proceed, and also considers the person riding a tricycle on the right side, thereby making sensible driving decisions. These qualitative results demonstrate our model's exceptional ability to understand complex scenarios and make suitable driving plans. More visualization of our model's output is shown in the Appendix~\ref{sec:appendix-vis}.

\vspace{2em}

\section{Onboard Deployment and Testing}
We deploy DriveVLM-Dual on an autonomous vehicle equipped with two OrinX processors, with a high-frequency end-to-end driving system on OrinX-1 and our DriveVLM on OrinX-2. These two systems operate and collaborate asynchronously. Furthermore, we optimize the DriveVLM and achieve an average inference speed of 410 ms on OrinX. The video demonstration of this dual-system deployment can be found at the following link: \url{https://www.youtube.com/watch?v=MMCO0TLMT74}. 

In this section, we present several comparative experiments, which will provide insights for practical deployment on edge devices. Notice that all the VLMs are pre-trained and fine-tuned on common datasets as well as our proprietary datasets specifically designed for autonomous driving.

\begin{table}[t]
\centering
\resizebox{0.99\columnwidth}{!}{
\begin{tabular}{l|cccccccc}
\toprule
\multirow{2}{*}{Base LLM} & \multirow{2}{*}{Avg.} & MMMU~\cite{mmmu} & SEEDBench~\cite{seedbench} & RefCOCO~\cite{refcoco} & SUP-AD & Drivelm-QA~\cite{drivelm} & Drivelm-Grounding~\cite{drivelm} & RealworldQA~\cite{grok-1.5v} \\
    &                 & 5\%  & 20\%      & 15\%    & 15\% & 7.5\%      & 7.5\%             & 15\%           \\
\midrule
MobileLLaMA1.4B~\cite{mobilevlm} & 0.457 & 0.331 & 0.590 & 0.421 & 0.520 & 0.686 & 0.735 & 0.501 \\
Qwen-1.8B~\cite{qwen1.5}  & 0.477 & 0.340 & 0.622 & 0.492 & 0.523 & 0.680 & 0.725 & 0.518 \\
Gemma-2B~\cite{gemma}  & 0.439 & 0.345 & 0.571 & 0.330 & 0.510 & 0.680 & 0.721 & 0.507 \\
MiniCPM-2.4B~\cite{minicpm} & 0.482 & 0.379 & 0.640 & 0.444 & 0.539 & 0.676 & 0.717 & 0.553 \\
MobileLLaMA2.7B~\cite{mobilevlm} & 0.496 & 0.348 & 0.635 & 0.557 & 0.546 & 0.683 & 0.725 & 0.536 \\
Phi3-3.8B~\cite{phi3}  & 0.538 & 0.435 & 0.688 & 0.608 & 0.604 & 0.697 & 0.743 & 0.592 \\
Qwen-4B~\cite{qwen1.5} & 0.511  & 0.366 & 0.671 & 0.603 & 0.515 & 0.681 & 0.735 & 0.562 \\
Qwen-4B$^{\ast}$ & 0.529 & 0.373 & 0.684 & 0.624 & 0.596 & 0.699 & 0.738 & 0.553 \\
\bottomrule
\end{tabular}
}
\vspace{0.5em}
\caption{Performance of different LLMs on various datasets using the LLAVA-1.5~\cite{llava-1.5} architecture with ViT-L-336~\cite{vit} as the image encoder. Note that $^{\ast}$ indicates using SigLIP-L-384~\cite{siglip} as the image encoder. Compared to general capabilities, we are more concerned with the performance of VLMs in abilities that are more relevant to autonomous driving. Therefore, we set different score weights for various benchmarks.}
\label{tab:sup-llm-performance}
\end{table}

\begin{table}[t]
\centering
\resizebox{0.99\columnwidth}{!}{
\begin{tabular}{l | c c c c c c c c c c}
\toprule
LLM & Promt Length(toks) & Prefill latency (s) & Prefill (tok/s) & Decode (tok/s) & Output (toks) & Decode latency (s) & Model Size (GB) & Layer Num & Head Size & Vocab Size \\
\midrule
Gemma-2B~\cite{gemma}         & 1063 & 0.95 & 1121 & 40.9  & 59 & 1.44 & 4.7 & 18 & 256 & 256000 \\
Phi3-4k~\cite{phi3}          & 1045 & 1.30  & 797.3 & 49.0  & 59 & 1.20  & 7.2 & 32 & 96  & 32064  \\
MobileLLaMa-2.7B~\cite{mobilevlm} & 1047 & 0.92 & 1134 & 61.7 & 59 & 0.96 & 5.0 & 32 & 80  & 32000  \\
MobileLLaMa-1.4B~\cite{mobilevlm} & 1047 & 0.23 & 4634 & 117.4& 59 & 0.50  & 2.5 & 24 & 128 & 32000  \\   
Qwen4B~\cite{qwen1.5}           & 1078 & 0.57 & 1882 & 44.5 & 59 & 1.33 & 7.5 & 40 & 128 & 151936 \\
Qwen1.8B~\cite{qwen1.5}            & 1078 & 0.23 & 4709 & 79.6 & 59 & 0.74 & 3.7 & 24 & 128 & 151936 \\
\bottomrule
\end{tabular}
}
\vspace{0.5em}
\caption{Inference performance of different LLMs after quantization and deployment on an OrinX chip. The Qwen series achieved the best performance.}
\label{tab:sup-llm-speed}
\end{table}

\begin{table}[b]
\centering
\resizebox{0.99\columnwidth}{!}{
\begin{tabular}{l c l | c c c c c c c c c}
\toprule
\multirow{2}{*}{Backbone} & \multirow{2}{*}{Token Length} & \multirow{2}{*}{Method} & \multirow{2}{*}{Weighted Total} & MMMU~\cite{mmmu} & SEEDBench~\cite{seedbench} & RefCOCO~\cite{refcoco} & SUP-AD & Drivelm-QA~\cite{drivelm} & Drivelm-Grounding~\cite{drivelm} & RealworldQA~\cite{grok-1.5v} & PointVQA \\
       &              &                   &                & 5\%  & 20\%      & 15\%    & 15\% & 7.5\%      & 7.5\%             & 15\%           & 15\%     \\
\midrule
ViT-L-336~\cite{vit} & - & - & 0.421 & 0.368 & 0.657 & 0.502 & 0.553 & 0.688 & 0.742 & 0.541 & 0.583 \\
ViT-L-336 & - & Fixed 1x2 + CA-256 & 0.413 & 0.388 & 0.62 & 0.327 & 0.543 & 0.699 & 0.733 & 0.536 & 0.563 \\
ViT-L-336 & - & Fixed 2x2 + CA-256 & 0.405 & 0.376 & 0.618 & 0.306 & 0.518 & 0.692 & 0.731 & 0.523 & 0.594 \\
ViT-L-448 & - & Fixed 1x2 + CA-256 & 0.383 & 0.391 & 0.542 & 0.184 & 0.528 & 0.689 & 0.718 & 0.470 & 0.562 \\
ViT-L-336 & - & Fixed 1x2 + S2 & 0.368 & 0.383 & 0.511 & 0.198 & 0.469 & 0.689 & 0.718 & 0.469 & 0.314 \\
ViT-L-336 & - & Fixed 1x2 + S2 & 0.368 & 0.381 & 0.537 & 0.207 & 0.544 & 0.698 & 0.729 & 0.460 & 0.554 \\
ViT-L-336 & - & DM-4 + CA-256 & 0.409 & 0.377 & 0.610 & 0.277 & 0.544 & 0.698 & 0.726 & 0.525 & 0.554 \\
ViT-L-336 & - & DM-6 + CA-256 & 0.381 & 0.381 & 0.500 & 0.162 & 0.523 & 0.700 & 0.729 & 0.466 & 0.584 \\
ViT-L-336 & - & DM-4 + PT & 0.426 & 0.376 & 0.653 & 0.557 & 0.585 & 0.700 & 0.743 & 0.540 & 0.598 \\
ViT-L-336 & - & DM-4 + SM + LT & 0.413 & 0.403 & 0.617 & 0.293 & 0.550 & 0.699 & 0.737 & 0.527 & 0.532 \\
ViT-L-336 & - & DM-4 + SM + AAP + LT & 0.406 & 0.384 & 0.610 & 0.273 & 0.518 & 0.68 & 0.726 & 0.520 & 0.533 \\
ViT-L-336 & - & DM-4 + SM + CD + LT & 0.409 & 0.388 & 0.630 & 0.480 & 0.515 & 0.700 & 0.734 & 0.524 & 0.577 \\
SigLIP-L-384~\cite{siglip} & 576 & - & 0.432 & 0.389 & 0.631 & 0.615 & 0.624 & 0.707 & 0.749 & 0.556 & 0.560 \\
SigLIP-L-768 & 576 & - & 0.438 & 0.377 & 0.642 & 0.580 & 0.638 & 0.712 & 0.764 & 0.561 & 0.556 \\
SigLIP-L-1152 & 576 & - & 0.436 & 0.377 & 0.637 & 0.595 & 0.628 & 0.715 & 0.763 & 0.571 & 0.564 \\
SigLIP-L-384-768 & 576 & PE & 0.434 & 0.367 & 0.632 & 0.581 & 0.631 & 0.716 & 0.762 & 0.556 & 0.554 \\
SigLIP-L-512-960 & 480 & PE & 0.442 & 0.369 & 0.640 & 0.557 & 0.650 & 0.719 & 0.762 & 0.579 & 0.568 \\
\bottomrule
\end{tabular}
}
\vspace{0.5em}
\caption{Performance of different methods for scaling ViT's original input resolution to higher resolution. ``CA'' stands for cross-attention, ``DM'' is the abbreviation for Dynamic Max, ``PT'' indicates the use of patch end token, ``SM'' stands for Spatial Merge, ``LT'' indicates the use of line end token, ``AAP'' is the abbreviation for Adaptive Average Pooling, and ``CD'' stands for Convolutional Downsampling. ``PE'' represents Position Embedding Interpolation. Among these methods, applying PE interpolation to SigLIP-L-384 and modifying it to take 768-resolution images as input achieves a good trade-off between inference speed and performance.
}
\label{tab:sup-scale-up-image}
\end{table}

\paragraph{Base LLM}
 Due to the limited memory and bandwidth of the vehicle's hardware, we cannot use overly large LLMs to maintain real-time inference. Therefore, we chose models with fewer than 4 billion parameters. As shown in Table~\ref{tab:sup-llm-performance} and \ref{tab:sup-llm-speed}, our experiments revealed that on the Orin architecture, the ``wide and shallow'' Qwen series (wider and fewer layers) models outperform ``narrow and deep'' models (narrower and more layers) in inference speed. 

\paragraph{Visual Encoder}
High-resolution images are essential for fine-grained visual understanding in autonomous driving. As shown in Table~\ref{tab:sup-scale-up-image}, compared to the basic ViT model used as a visual encoder, we explored several options, including different GridPatch strategies and PE (Position Embedding) interpolation. Ultimately, for real-time inference, we selected the simpler SigLIP-L-384 model with PE interpolation, achieving high-resolution input through original 384-resolution PE interpolation and fine-tuning parameters with additional convolution layers.

\begin{table}[t]
\centering
\resizebox{0.99\columnwidth}{!}{
\begin{tabular}{l | c c c c c c c c c c c c c}
\toprule
Projector & Origin & Compressed & Output(ms) & Prefill(ms) & Avg. & MMMU~\cite{mmmu} & SEEDV2~\cite{seedbench} & BDD~\cite{bddx} & Drivelm-QA~\cite{drivelm} & Drivelm-grounding~\cite{drivelm} & RealworldQA~\cite{grok-1.5v} & RefCOCO~\cite{refcoco} & SUP-AD \\
\midrule
MLP (Baseline) & 576 & 576 & 666.60 & 707.93 & 56.27 & 37.90 & 64.00 & 53.90 & 67.60 & 71.70 & 55.30 & 44.40 & 53.90 \\
LDPNetV2~\cite{mobilevlm} & 576 & 576 & 661.70 & 707.42 & 50.53 & 37.67 & 56.71 & 53.36 & 68.90 & 73.20 & 48.30 & 21.29 & 54.70 \\
Perceiver Resampler~\cite{flamingo} & 576 & 576 & 637.20 & 666.20 & 49.71 & 39.93 & 58.84 & 52.21 & 68.37 & 71.90 & 49.61 & 19.01 & 48.70 \\
Pixel shuffle & 576 & 256 & 610.70 & 655.92 & 52.02 & 40.42 & 60.20 & 55.19 & 68.72 & 71.52 & 51.44 & 28.23 & 48.40 \\
LDPNetV2 & 576 & 256 & 605.32 & 652.79 & 54.73 & 38.93 & 62.19 & 54.82 & 68.57 & 72.78 & 50.59 & 34.32 & 59.30 \\
Pixel shuffle & 576 & 144 & 604.98 & 645.61 & 49.40 & 38.98 & 61.63 & 56.06 & 69.18 & 73.18 & 51.05 & 30.67 & 58.80 \\
LDPNetV2 & 576 & 144 & 597.15 & 646.77 & 55.56 & 38.93 & 62.63 & 56.23 & 68.93 & 72.48 & 50.59 & 33.14 & 59.30 \\
LDPNetV2 & 576 & 64 & 616.20 & 645.18 & 56.24 & 39.40 & 61.80 & 54.60 & 68.80 & 72.60 & 51.00 & 41.88 & 61.20 \\
\bottomrule
\end{tabular}
}
\vspace{0.5em}
\caption{Performance of different methods for visual token compression. Using LDPNetV2 to compress the original tokens to 75\% of the original token count, achieves the best trade-off between performance and speed.}
\label{tab:sup-token-compression}
\end{table}

\paragraph{Visual Token Compression}
To address the increased computational load from high-resolution images, we implemented LDPNetv2~\cite{mobilevlm} to reduce the number of image tokens by 75\% without compromising performance, as shown in Table~\ref{tab:sup-token-compression}. Additionally, we enhanced performance by replacing the average pooling layer with a convolution layer in LDPNetv2. 

\paragraph{Video Input}
In autonomous driving scenarios, more temporal context is needed for accurately assessing object motion changes. We employ a short-term memory-bank~\cite{memory-bank} strategy, temporarily storing visual features from historical frames in a feature queue. Only the features from the current moment are extracted and fused with multiple historical frames before being projected into the LLM. Besides basic spatiotemporal pooling, we added SE~\cite{se-net} blocks to perform a weighted fusion of multiple temporal frames.

\begin{table}[!t]
\centering
\resizebox{0.99\columnwidth}{!}{
\begin{tabular}{l | c c c c c}
\toprule
     & \multirow{2}{*}{Base + q4f16\_1} & \multirow{2}{*}{Eagle~\cite{eagle} + q4f16\_1} & \multirow{2}{*}{Medusa~\cite{medusa} + q4f16\_ft} & \multirow{2}{*}{Eagle + q4f16\_ft} & Eagle + q4f16\_ft  \\
 & & & & & + Shrink Vocab Size(1024) \\
\midrule
Quant Type & q4f16\_1 & q4f16\_1 & q4f16\_ft & q4f16\_ft & q4f16\_ft \\
Input Size & (384, 960) & (384, 960) & (384, 960) & (384, 960) & (384, 960) \\
Prefill Tokens & 604 & 604 & 613 & 604 & 613 \\
Output Tokens  & 37 & 39 & 41 & 41 & 41 \\
Prefill Speed (tok/s) & 1818 & 1793 & 2160.64 & 2230 & 2175.35 \\
Decode Speed (tok/s) & 109 & 172 & 232.34 & 295.04 & 518.3 \\
Prefill Latency (s) & 0.332 & 0.340 & 0.284 & 0.276 & 0.274 \\
Decode Latency (s) & 0.328 & 0.216 & 0.176 & 0.130 & 0.071 \\
Acceleration ratio & 1 & 1.57 & 2.17 & 2.7 & 4.33 \\
\bottomrule
\end{tabular}
}
\vspace{0.5em}
\caption{Performance of different speculative sampling methods. ``Shrink Vocab Size'' means we reduce the vocabulary to the 1024 most frequently used words. ``q4f16\_1'' is a 4-bit quantization method using a 16-bit floating-point representation for efficient model compression, while ``q4f16\_ft'' includes subsequent fine-tuning to enhance performance post-quantization.}
\label{tab:sup-decode}
\end{table}

\paragraph{Speculative Sampling}
Speculative Sampling is used to accelerate inference by preemptively generating likely outputs. This approach reduces the latency of generating predictions, achieving a significant speedup without substantial loss in accuracy. As shown in Table~\ref{tab:sup-decode}, we test two speculative sampling methods: Medusa~\cite{medusa} and Eagle~\cite{eagle} with our inference framework designed specifically for the OrinX chip. Eagle achieved a 2.7 \(\times\) speedup in decode latency compared to Medusa's 2.17 \(\times\), making real-time vehicle deployment feasible.

\section{Conclusion}
In summary, we introduce DriveVLM and DriveVLM-Dual. DriveVLM leverages VLMs, significantly progressing in interpreting complex driving environments. DriveVLM-Dual further enhances these capabilities by synergizing existing 3D perception and planning approaches, effectively addressing the spatial reasoning and computational challenges inherent in VLMs. Moreover, we define a scene understanding for planning task for autonomous driving, together with evaluation metrics and dataset construction protocol.
DriveVLM and DriveVLM-Dual have surpassed the state-of-the-art methods on the public and our benchmarks, especially in handling intricate and dynamic scenarios. Finally, we have verified the effectiveness of DriveVLM-Dual through onboard deployment and testing on a production vehicle.
\vspace{2em}
\acknowledgments{
We thank Chenxu Hu at Tsinghua University for the help on paper writing. We thank Xu Bian, Hongkun Chen, Sui Cong, Chengze Guan, Mingyu Guo, Yue Jiang, Qi Jiang, Pengfei Ji, Wei Xiao, Dafeng Wei, Zijian Wang, Zhao Yang, Chenglong Zhao, Simeng Zhao, and Jian Zhou at Li Auto for their efforts in the experiments related to onboard deployment.}



\bibliography{egbib}

\clearpage

\appendix
\title{Appendix}
\section{SUP-AD Dataset}
\label{sec:appendix-dataset}

\subsection{Meta-actions}
\begin{figure}[!t]
    \centering
    \includegraphics[width=0.99\linewidth]{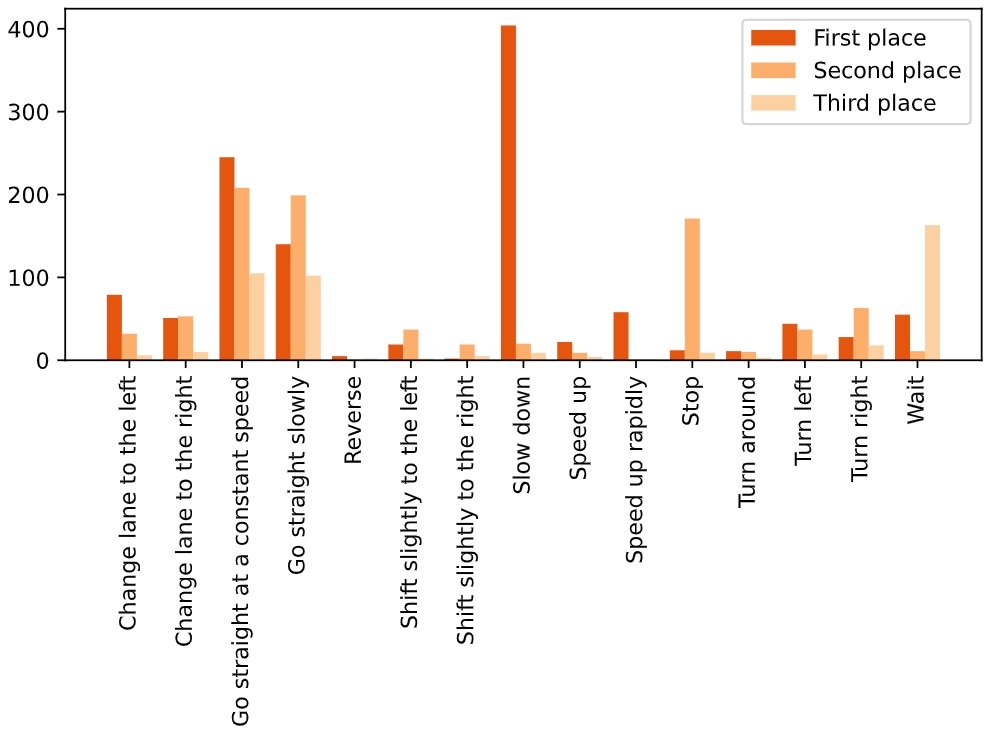}
    \caption{\textbf{Distribution of each meta action being the first, second, and third place of the meta action sequence, respectively.}}
    \label{fig:meta-action-stat}
    \vspace{6em}
\end{figure}

\begin{figure}[!h]
    \centering
    \includegraphics[width=0.99\linewidth]{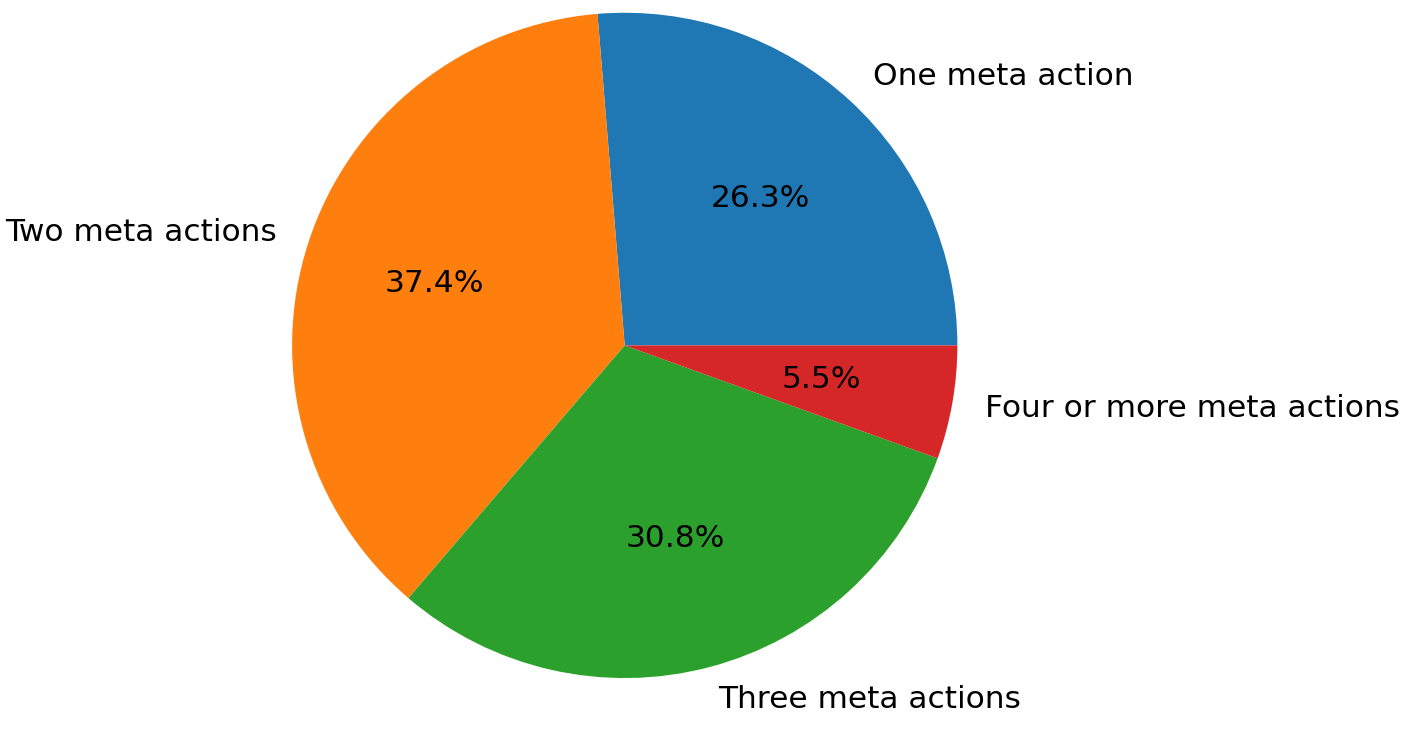}
    \caption{\textbf{Distribution of the length of meta actions per scene.}}
    \label{fig:stat-meta-action-length}
\end{figure}

\paragraph{Meta-action statistics.}
We use the meta-action sequence to formally represent the driving strategy. Meta actions are classified into 17 categories. We show the distribution of each meta-action being the first/second/third place in the meta-action sequence, as shown in Figure~\ref{fig:meta-action-stat}. It indicates that the meta-actions are quite diverse in the SUP-AD dataset. We also show the distribution of the length of meta-actions per scene in Figure~\ref{fig:stat-meta-action-length}. Most scenes contain two or three meta-actions, and a few scenes with complex driving strategies contain four or more meta-actions.

\paragraph{Annotation of meta-actions.}
The meta-action sequence for each driving scene is manually annotated based on the actual driving strategy in the future frames. These meta-actions are designed to encompass a complete driving strategy and are structured to be consistent with the future trajectory of the ego vehicle. They can be divided into three primary classes:
\begin{enumerate}
    \item \textbf{Speed-control actions.} Discerned from acceleration and braking signals within the ego state data, these actions include
    These actions can be discerned from acceleration and braking signals within the ego state data. They include \textit{speed up}, \textit{slow down}, \textit{slow down rapidly}, \textit{go straight slowly}, \textit{go straight at a constant speed}, \textit{stop}, \textit{wait}, and \textit{reverse}.
    \item \textbf{Turning actions.} Deduced from steering wheel signals, these actions consist of \textit{turn left}, \textit{turn right}, and \textit{turn around}.
    \item \textbf{Lane-control actions.} Encompassing lane selection decisions, these actions are derived from a combination of steering wheel signals and either map or perception data. They involve \textit{change lane to the left}, \textit{change lane to the right}, \textit{shift slightly to the left}, and \textit{shift slightly to the right}.
\end{enumerate}

\begin{figure*}[!h]
    \centering
    \includegraphics[width=0.97\linewidth]{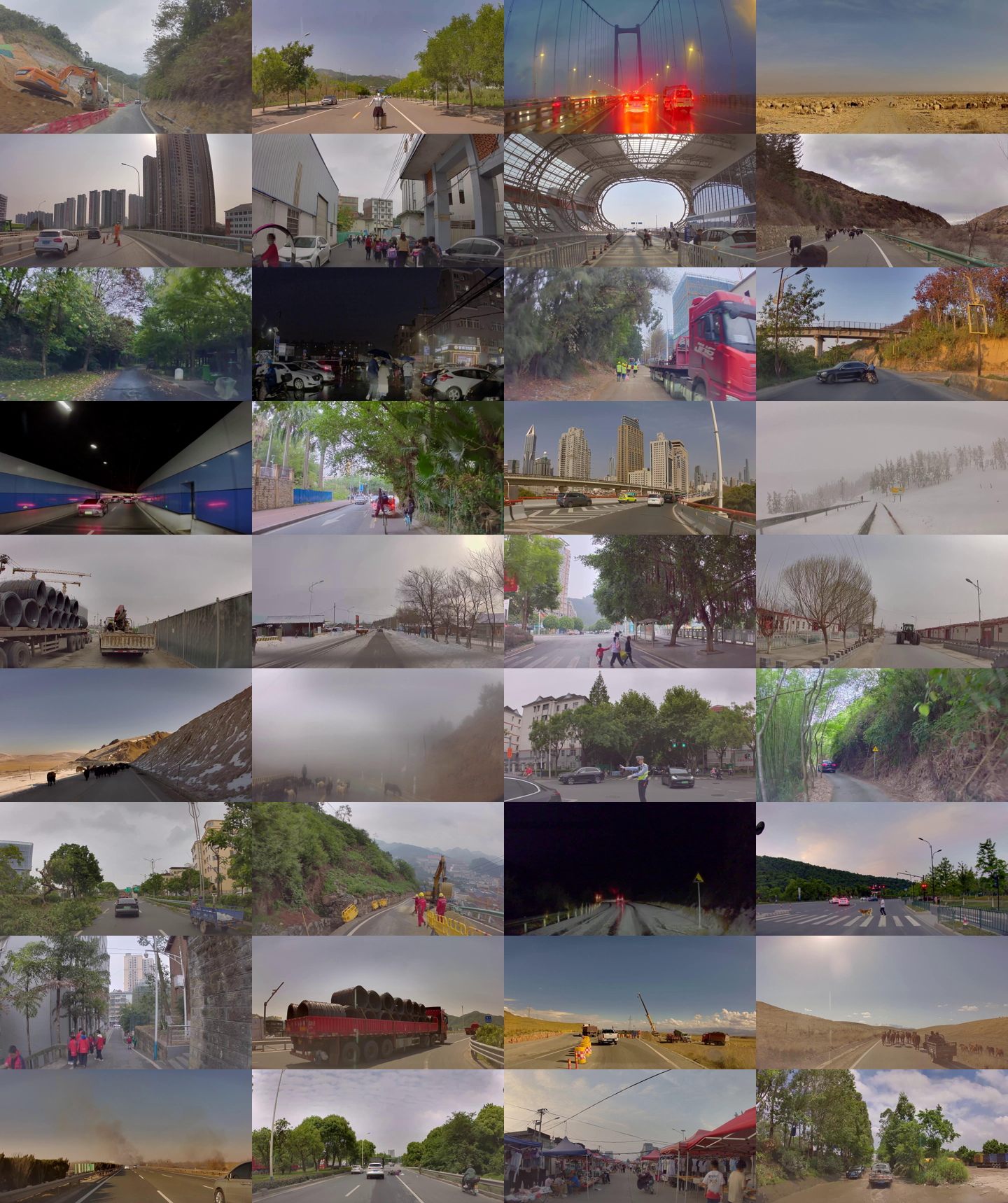}
    \vspace{20pt}
    \caption{\textbf{Diverse driving scenarios in the SUP-AD dataset.}}
    \label{fig:overview}
\end{figure*}



\subsection{Scenario Categories}
The SUP-AD dataset is comprised of 1,000 video clips of driving scenarios. As illustrated in Figure~\ref{fig:overview}, it encompasses a wide range of driving scenarios, spanning over 40 categories. Below are explanations for some of the scenarios:

\textbf{AEB Data}: Automatic Emergency Braking (AEB) data.

\textbf{Road Construction}: A temporary work zone with caution signs, barriers, and construction equipment ahead.

\textbf{Close-range Cut-ins}: A sudden intrusion into the lane of the ego vehicle by another vehicle.

\textbf{Roundabout}: A type of traffic intersection where vehicles travel in a continuous loop.

\textbf{Animals Crossing Road}: Animals crossing the road in front of the ego vehicle.

\textbf{Braking}: Brake is pressed by human driver of the ego vehicle.

\textbf{Traffic Police Officers}: Traffic police officers managing and guiding traffic.

\textbf{Blocking Traffic Lights}: A massive vehicle obscuring the visibility of the traffic signal.

\textbf{Cutting into Other Vehicle}: Intruding into the lane of another vehicle ahead.

\textbf{Ramp}: A curved roadway that connects the main road to the branch road in highway.

\textbf{Debris on the Road}: Road with different kinds of debris.

\textbf{Narrow Roads}: Narrow roads that require cautious navigation.

\textbf{Pedestrians Popping Out}: Pedestrians popping out in front of the ego vehicle, requiring slowing down or braking.

\textbf{People on Bus Posters}: Buses with posters, which may interfere the perception system.

\textbf{Merging into High Speed}: Driving from a low-speed road into a high-speed road, requiring speeding up.

\textbf{Barrier Gate}: Barrier gate that can be raised obstructing the road.

\textbf{Fallen Trees}: Fallen trees on the road, requiring cautious navigation to avoid potential hazards.

\textbf{Complex Environments}: Complex driving environments that requiring cautious navigation.

\textbf{Mixed Traffic}: A congested scenario where cars, pedestrians, and bicycles appear on the same or adjacent roadway.

\textbf{Crossing Rivers}: Crossing rivers by driving on the bridge.

\textbf{Screen}: Roads with screens on one side, which may interfere the perception system.

\textbf{Herds of Cattle and Sheep}: A rural road with herds of cattle and sheep, requiring careful driving to avoid causing distress to these animals.

\textbf{Vulnerable Road Users}: Road users which are more susceptible to injuries while using roads, such as pedestrians, cyclists, and motorcyclists.

\textbf{Road with Gallet}: A dusty road with gallet scattered across the surface.

The remaining scenario categories are: Motorcycles and Trikes, Intersection, People carrying Umbrella, Vehicles Carrying Cars, Vehicles Carrying Branches, Vehicles with Pipes, Strollers, Children, Tunnel, Down Ramp, Sidewalk Stalls, Rainy Day, Crossing Train Tracks, Unprotected U-turns, Snowfall, Large Vehicles Invading, Falling Leaves, Fireworks, Water Sprinklers, Potholes, Overturned Motorcycles, Self-ignition and Fire, Kites, Agricultural Machinery.


\subsection{Annotation Examples}
\begin{figure*}[t]
    \centering
    \includegraphics[width=0.98\linewidth]{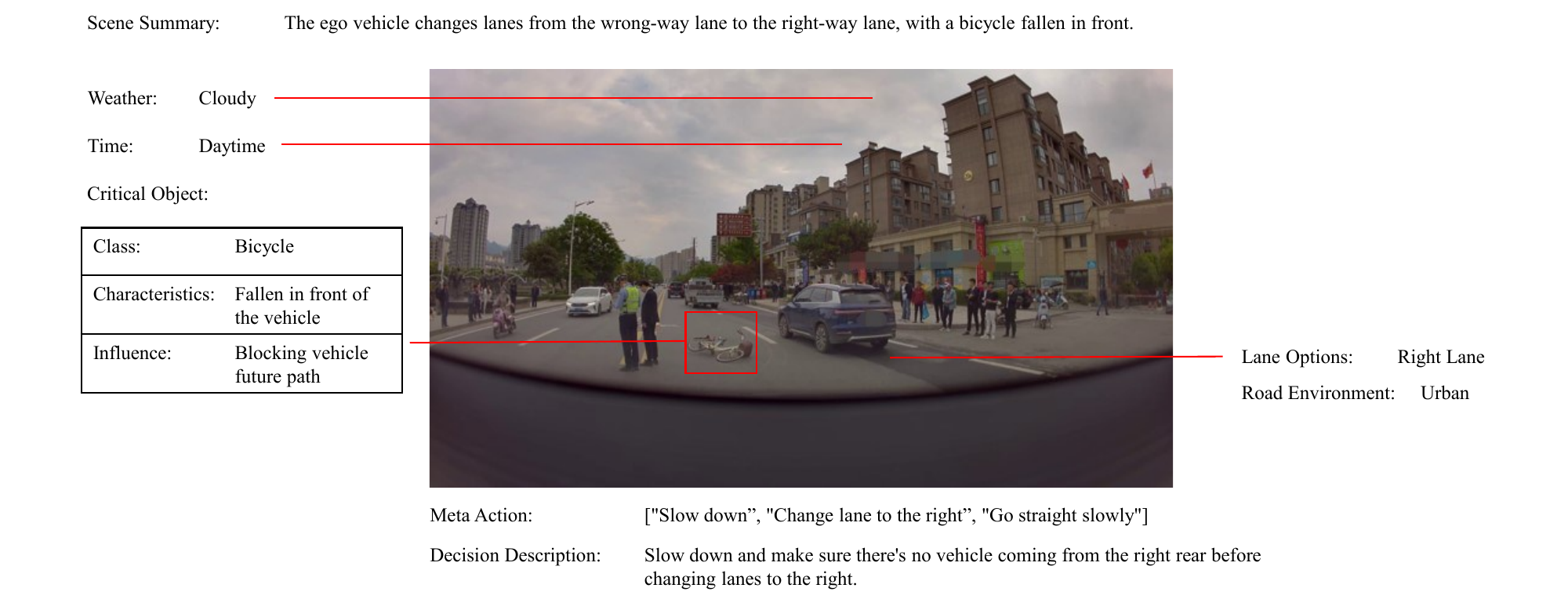}
    \caption{\textbf{An example of overturned bicycles and motorcycles in the SUP-AD dataset.} A bicycle has fallen in front of the ego vehicle, requiring the ego vehicle to change lanes.}
    \label{fig:example2}
\end{figure*}

\begin{figure*}[!ht]
    \centering
    \includegraphics[width=0.99\linewidth]{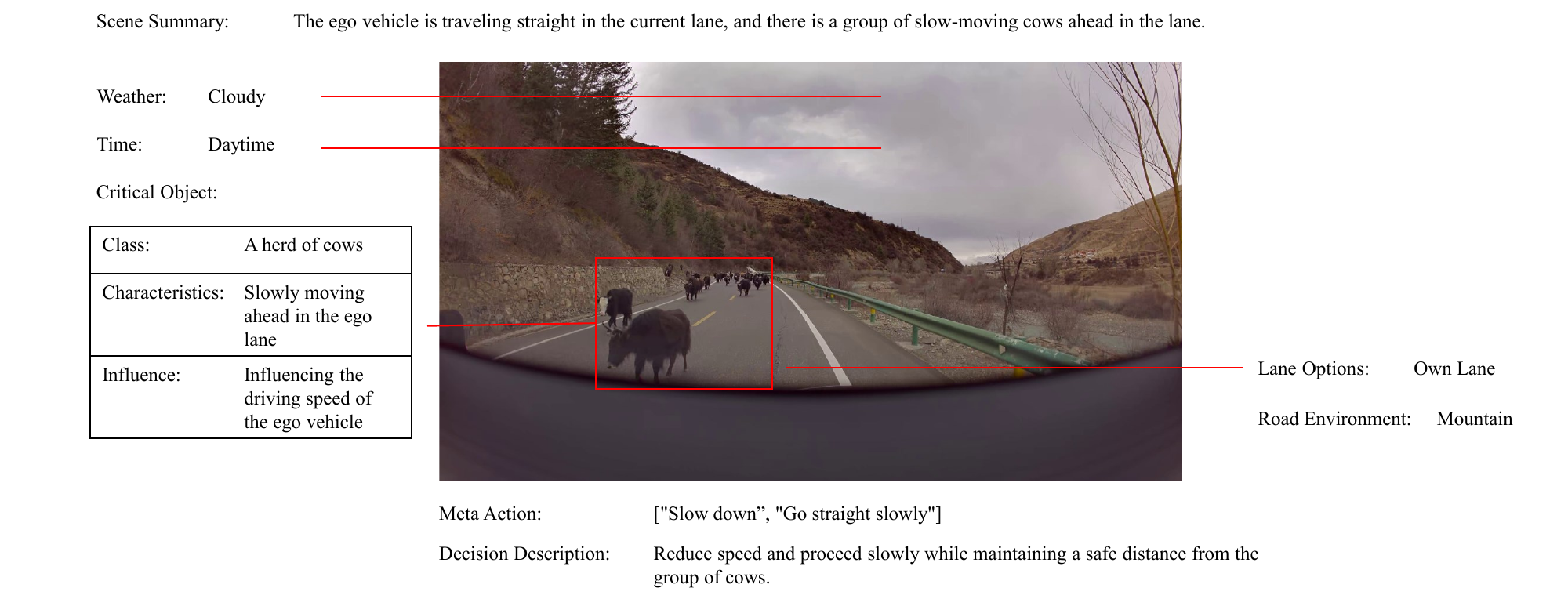}
    \caption{\textbf{An example of herds of cattle and sheep in the SUP-AD dataset.} A group of cattle move slowly in front of the ego vehicle, requiring the ego vehicle to proceed slowly and maintain a safe distance from the cattle.}
    \label{fig:example3}
\end{figure*}

\begin{figure*}[!ht]
    \centering
    \includegraphics[width=0.99\linewidth]{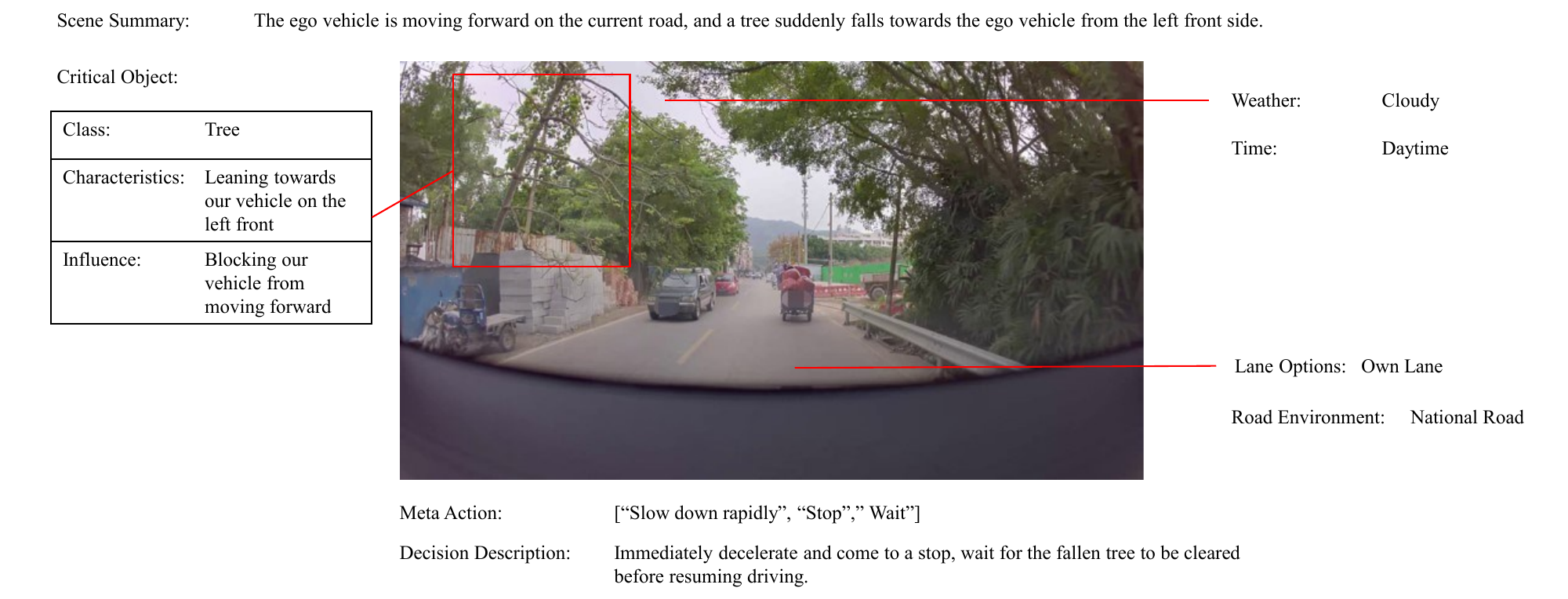}
    \caption{An example of collapsed trees in the SUP-AD dataset. A tree suddenly falls towards the ego vehicle, requiring the ego vehicle to decelerate immediately.}
    \label{fig:example4}
\vspace{1em}
\end{figure*}

\begin{figure*}[!ht]
    \centering
    \includegraphics[width=0.99\linewidth]{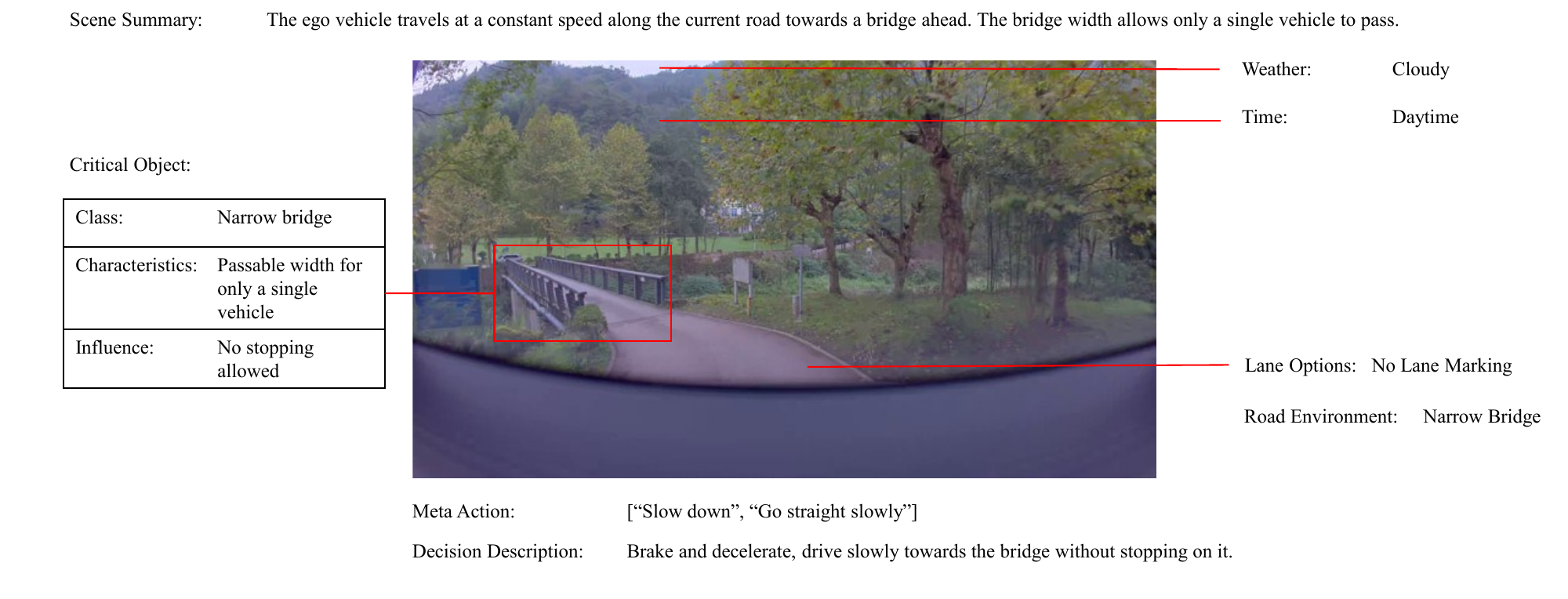}
    \caption{\textbf{An example of crossing rivers in the SUP-AD dataset.} The ego vehicle is going across a bridge of which width allows only a single vehicle to pass, requiring the ego vehicle to drive without stopping.
}
    \label{fig:example5}
\vspace{1em}
\end{figure*}

\begin{figure*}[!ht]
    \centering
    \includegraphics[width=0.99\linewidth]{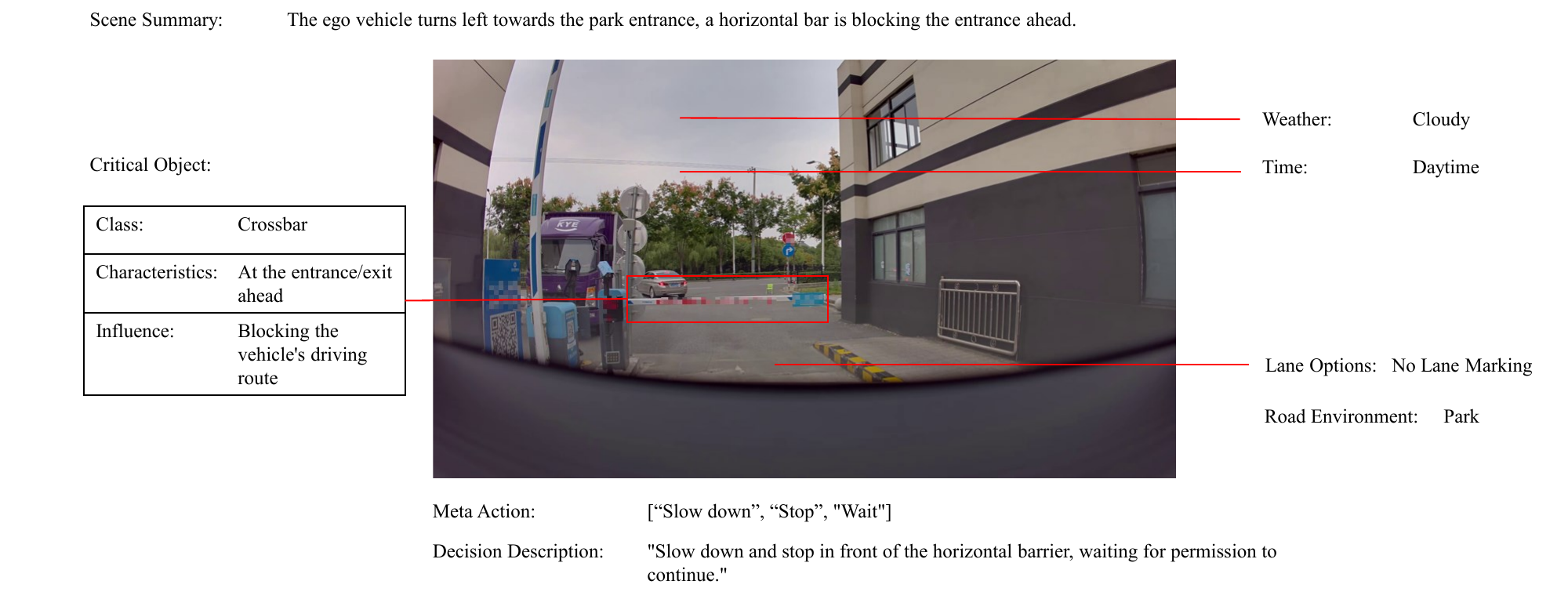}
    \caption{\textbf{An example of barrier gate in the SUP-AD dataset.} A horizontal barrier blocks the entrance of a park, requiring the ego vehicle to stop and wait for permission to continue.}
    \label{fig:example6}
\vspace{1em}
\end{figure*}

\begin{figure*}[!ht]
    \centering
    \includegraphics[width=0.99\linewidth]{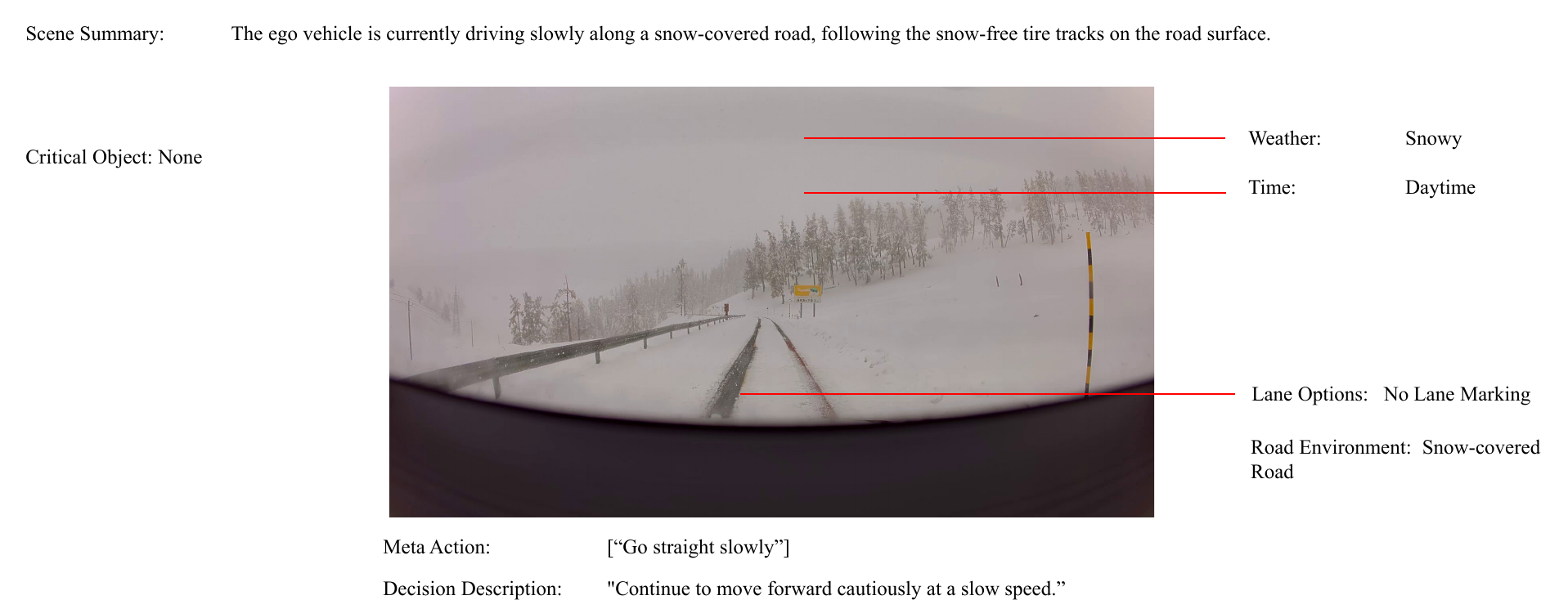}
    \caption{\textbf{An example of snowfall in the SUP-AD dataset.} Most of the road is covered by snow, requiring the ego vehicle to move forward cautiously by following the snow-free tire tracks.}
    \label{fig:example7}
\end{figure*}

We provide more examples of annotation contents in Figure~\ref{fig:example2},~\ref{fig:example3},~\ref{fig:example4},~\ref{fig:example5},~\ref{fig:example6}, and ~\ref{fig:example7}. The scenario categories of these examples are overturned bicycles and motorcycles, herds of cattle and sheep, collapsed trees, crossing rivers, barrier gate, and snowfall respectively. 

\section{Evaluation Method}
\label{sec:appendix-eval}
The ability of an autonomous driving system to accurately interpret driving scenes and make logical, suitable decisions is of paramount importance. As presented in this paper, the evaluation of VLMs in autonomous driving concentrates on two primary components: the evaluation of scene description/analysis and the evaluation of meta-actions.
\subsection{Scene Description/Analysis Evaluation}
In terms of scene description/analysis evaluation, the process of interpreting and articulating driving scenes is subject to inherent subjectivity, as there are numerous valid ways to express similar descriptions textually, which makes it difficult to effectively evaluate the scene description using a fixed metric. To overcome this challenge, we utilize GPT-4 to evaluate the similarity between the scene descriptions generated by the model and the manually annotated ground truth. Initially, we prompt GPT-4 to extract individual pieces of information from each scene description. Subsequently, we score and aggregate the results based on the matching status of each extracted piece of information.


The ground truth labels for scene descriptions encompass both environment descriptions and event summaries. Environmental condition description includes weather conditions, time conditions, road environment, and lane conditions. Event summaries are the characteristics and influence of critical objects. 
We employ GPT-4 to extract unique key information from both environment descriptions and event summaries. The extracted information is then compared and quantified. Each matched pair is assigned a score, which is estimated based on the extent of the matching, whether complete, partial, or absent. Instances of hallucinated information incur a penalty, detracting from the overall score. The aggregate of these scores constitutes the scene description score.

\begin{align}
\begin{split}
\text{Score} = \frac{1.0 \times n_{\text{matched}} + 0.5 \times n_{\text{partial}}}{n_{\text{gt}}} \\
- \frac{0.25 \times n_{\text{hallucination}}}{n_{\text{gt}}}
\end{split}
\end{align}


The prompt for GPT-4 in evaluating scene descriptions is carefully designed, as shown in Table~\ref{tab:description}. Initially, a role prompt is employed to establish as an intelligent and logical evaluator, possessing a comprehensive understanding of appropriate driving styles. This is followed by specifying the input format, which informs GPT-4 that its task involves comparing an output description with a ground truth description. This comparison is based on the extraction and analysis of key information from both descriptions. Lastly, the prompt outlines the criteria for scoring, as well as the format for the evaluation output, ensuring a structured and systematic approach to the evaluation process.
\begin{table*}[]
\scriptsize
\begin{tabular}{p{\textwidth}}
\toprule
\textbf{System Prompt:} \\
You are a smart and logical evaluator with extensive driving experience, and you will try your best to give reasonable and logical evaluation result. \\
\midrule
\textbf{Input Prompt:} \\
Given two driving scenario descriptions, one is the reference description, the other is the output description from a model. Please extract key information from the reference description and use it as comparison standards to compare the proximity of the output descriptions. Additionally, evaluate the completeness of each piece of information. Key information extracted can be classified into environmental conditions and critical events. Environmental condition information includes weather, time, road environment, and lane conditions. A critical event describes the event related to ego vehicle\'s driving decisions, which is usually given in the format of critical object + verb/adj + influence on ego vehicle. An example of a critical event is like this: A construction zone is blocking the ego lane and the right lane in front, so the car needs to go around to avoid it. Now I will give you the reference description and the output description. \\
The reference description: \\
Weather: Sunny. || Time: Day. || Road Environment: Urban. || Lane Options: Left Lane, Own Lane, Right Lane. || Ego Lane Position: Middle Lane. \\
Critical Events: There are police officers ahead inspecting passing vehicles. The police officer in front of your lane is signaling you to stop for inspection. There are also vehicles queuing in the left lane. The right lane is a merging lane and is currently separated, so you cannot change lanes to the right. \\
The output description: \\
Weather: Sunny. || Time: Day. \\
There is a vehicle driving in the distance in front of you. There are traffic police on the left and right, and the traffic police signal to stop. \\
Compare the two descriptions using extracted key information and output a similarities score on the scale from 0 to 1.0 and an explanation to describe the logic of calculating the similarity scale. Each matched key information scores 1.0, and each partially matched key information scores 0.5. Hallucinations will be penalized as well. If any key information exists in the output but not in the reference, a score of -0.25 will be assigned. The final score is the total score divided by the number of key information extracted in the reference description. For example, if the total score is 1.5, and the total number of key information in reference description is 3, then the final score would be 1.5/3 = 0.5. Please output the detailed extraction and evaluation process. \\
\midrule
\textbf{Response:} \\
Reference description key information: \\
1. Weather: Sunny (environmental conditions)  \\
2. Time: Day (environmental conditions)  \\
3. Road Environment: Urban (environmental conditions)  \\
4. Lane Options: Left Lane, Own Lane, Right Lane (environmental conditions)  \\
5. Ego Lane Position: Middle Lane (environmental conditions)  \\
6. Police officers ahead inspecting vehicles (critical event) \\
7. The police officer in front of ego vehicle signalling to stop (critical event) \\
8. Vehicles queuing in the left lane (critical event) \\
9. Cannot change to the right lane as it\'s an merging lane and currently separated (critical event) \\
Output description key information: \\
1. Weather: Sunny. \\
2. Time: Day. \\
3. Vehicle driving in the distance in front of driver \\
4. Traffic police on the left and right \\
5. Traffic police signal to stop \\
By comparing the output to the reference: \\
- Key information 1, 2, 6, and 7 are matched, each for 1.0 points(weather and time, police officers and their signal to stop) \\
- Key information 3 in the output is not mentioned in reference, deduction of -0.25 points. \\
No scores for:  \\
- Key information 3, 4, 5, 8, and 9. \\
Total score = (1.0 + 1.0 + 1.0 + 1.0 - 0.25) / 9 = 0.417. \\
\bottomrule
\caption{\textbf{Evaluation process of scene description by GPT-4.} Our proposed evaluation method is not only capable of extracting and evaluating information from structured scene descriptions but is also applicable to unstructured descriptions.}
\label{tab:description}
\end{tabular}
\end{table*}

\subsection{Meta-action Evaluation}

The evaluation process for the meta-action sequence must consider both the quantity and the sequential arrangement of the matched meta-actions. We employ dynamic programming to compare the model's output and the annotated ground truth. Our dynamic programming approach is similar to the method utilized in identifying the longest common subsequence, albeit with two supplementary considerations.


The first consideration acknowledges the unequal weighting of different meta-actions. For instance, certain meta actions such as ``\textit{Slow Down}", ``\textit{Wait}", and ``\textit{Go Straight Slowly}" exhibit a greater emphasis on attitude rather than action. The presence or absence of these actions from a meta-action sequence does not alter the basic semantic essence of driving decisions but rather modifies the driving strategy to be either more assertive or more cautious. For example, a meta action sequence of ``\textit{Slow Down \texttt{->} Stop \texttt{->} Wait}" conveys a similar driving decision as a sequence with only the meta action ``Stop". Consequently, these sequences should not incur a penalty comparable to other meta actions such as ``Turn Left" or ``Change Lane to the Right". Therefore, these are designated as ``conservative actions”, and a reduced penalty is applied when they do not match during sequence evaluation.

The second consideration addresses the potential semantic equality among different meta-action sequences. For example, the sequences ``\textit{Change Lane to the Left \texttt{->} Speed Up \texttt{->} Go Straight At a Constant Speed \texttt{->} Change Lane to the Right}" and ``\textit{Change Lane to the Left \texttt{->} Speed Up Rapidly \texttt{->} Go Straight At a Constant Speed \texttt{->} Change Lane to the Right}" might both represent valid approaches to overtaking a slow-speed vehicle ahead. Recognizing that different meta-action sequences might convey similar meanings, we initially use GPT-4 to generate variant sequences that have comparable semantic meanings, in addition to the unique ground truth meta-action sequence, as shown in Table~\ref{tab:alternative-actions}. In the subsequent sequence-matching phase of the evaluation, all these variations, together with the manually annotated ground truth, are taken into consideration. The highest-scoring matching is then adopted as the definitive score for the final decision evaluation.

\begin{table*}[ht]
\centering
\begin{tabular}{p{\textwidth}}
\toprule
\textbf{Prompt:} \\

Given a predefined set of action tokens and a sequence of reference driving actions, find all alternative sequences of actions that have similar meanings but choose slightly different action tokens from the predefined action tokens. \\
Predefined action tokens: [Speed up, Slow down, Speed up rapidly, Slow down rapidly, Go straight slowly, Go straight at a constant speed, Turn left, Turn right, Change lane to the left, Change lane to the right, Shift slightly to the left, Shift slightly to the right, Stop, Wait, Turn around, Reverse]. \\
For example, if the reference driving action sequence is: Slow down \texttt{->} Shift slightly to the right \texttt{->} Go straight at a constant speed, alternative action sequences with similar meanings could be: \\
1. Slow down \texttt{->} Change lane to the right \texttt{->} Go straight at a constant speed; \\
2. Slow down rapidly \texttt{->} Shift slightly to the right \texttt{->} Go straight at a constant speed; \\
3. Slow down \texttt{->} Change lane to the right \texttt{->} Go straight slowly;  \\
4. Slow down \texttt{->} Shift slightly to the right \texttt{->} Go straight slowly. \\
Now the reference action sequence is: Slow down \texttt{->} Shift slightly to the right \texttt{->} Go straight at a constant speed. What are the alternative action sequences? \\

\midrule
\textbf{Response:} \\
Given the reference sequence: "Slow down \texttt{->} Shift slightly to the right \texttt{->} Go straight at a constant speed." Alternative action sequences with similar meanings based on the provided action tokens are: 
1. Slow down \texttt{->} Change lane to the right \texttt{->} Go straight at a constant speed. \\
2. Slow down rapidly \texttt{->} Shift slightly to the right \texttt{->} Go straight at a constant speed. \\
3. Slow down \texttt{->} Change lane to the right \texttt{->} Go straight slowly. \\
4. Slow down \texttt{->} Shift slightly to the right \texttt{->} Go straight slowly. You've provided this example in your question. \\
Based on the tokens and the context, these are indeed appropriate alternative sequences. \\
\bottomrule
\end{tabular}
\vspace{1em}
\caption{\textbf{Example of generating alternative action sequences by GPT-4.}}

\label{tab:alternative-actions}
\end{table*}

The state of dynamic programming is saved in a 2D matrix, wherein each row corresponds to a meta action in the ground truth action sequence, and each column corresponds to a meta action in the model output action sequence, noted as $S^{\,r,\,c}$. The dynamic programming initiates recursive calculations beginning from the first meta action of both sequences. Each element of the 2D matrix encompasses the optimal total score at the current matching position, as well as the preceding matching condition that yielded the optimal matching. In our dynamic programming algorithm, three transition equations govern distinct cases: \(S_{\text{missing}}\) for missing matching, \(S_{\text{redundant}}\) for redundant matching, and \(S_{\text{matching}}\) for successful matching. Successful matching occurs when the meta action is identical at the $r^{th}$ position in the reference sequence and the $c^{th}$ position in the model-generated sequence. In the case of missing matching, the meta action at the $r^{th}$ position in the reference sequence is unmatched, prompting a comparison with the ${r-1}^{th}$ position in the reference sequence and the $c^{th}$ position in the model-generated sequence. Conversely, redundant matching implies that the meta action at the $c^{th}$ position in the model-generated sequence is unmatched, leading to further examination of the $r^{th}$  position in the reference and the ${c-1}^{th}$ position in the model-generated sequence. The transformation equations for these cases are as follows:
\begin{align}
\begin{split}
    S_{\text{missing}}^{\,r,\,c} &= S^{\,r-1,\,c} - p_{\text{missing}}, \\
    S_{\text{redundant}}^{\,r,\,c} &= S^{\,r,\,c-1} - p_{\text{redundant}}, \\
    S_{\text{matching}}^{\,r,\,c} &= S^{\,r-1,\,c-1} + s_{\text{matching}}, \\
    S^{\,r,\,c} &= \max(S_{\text{missing}}^{\,r,\,c}, S_{\text{redundant}}^{\,r,\,c}, S_{\text{matching}}^{\,r,\,c}),
\end{split}
\end{align}

where \(s_{\text{matching}} = 1.0\) represents the reward score after a successful matching. If an action considered missing or redundant is classified as a conservative action, the penalties \(p_{\text{missing}}\) and \(p_{\text{redundant}}\) are quantified as half of \(s_{\text{matching}}\), i.e., 0.5. Conversely, if an action is not conservative, both penalties are assigned the same magnitude as \(s_{\text{matching}}\), i.e., 1.0. This approach is based on the premise that omitting a crucial meta action or inaccurately introducing a non-existent one equally hampers the effectiveness of the action sequence. The final score $Score_\text{{action}}$ should be divided by the length of the selected reference meta-action sequence, formulated as follow:
\begin{align}
Score_{\,\text{action}} = \frac{S^{\,r,\,c}}{N_{r}}
\end{align}

\section{Qualitative Results}
\label{sec:appendix-vis}
\begin{figure*}[!ht]
    \centering
    \includegraphics[width=0.99\linewidth]{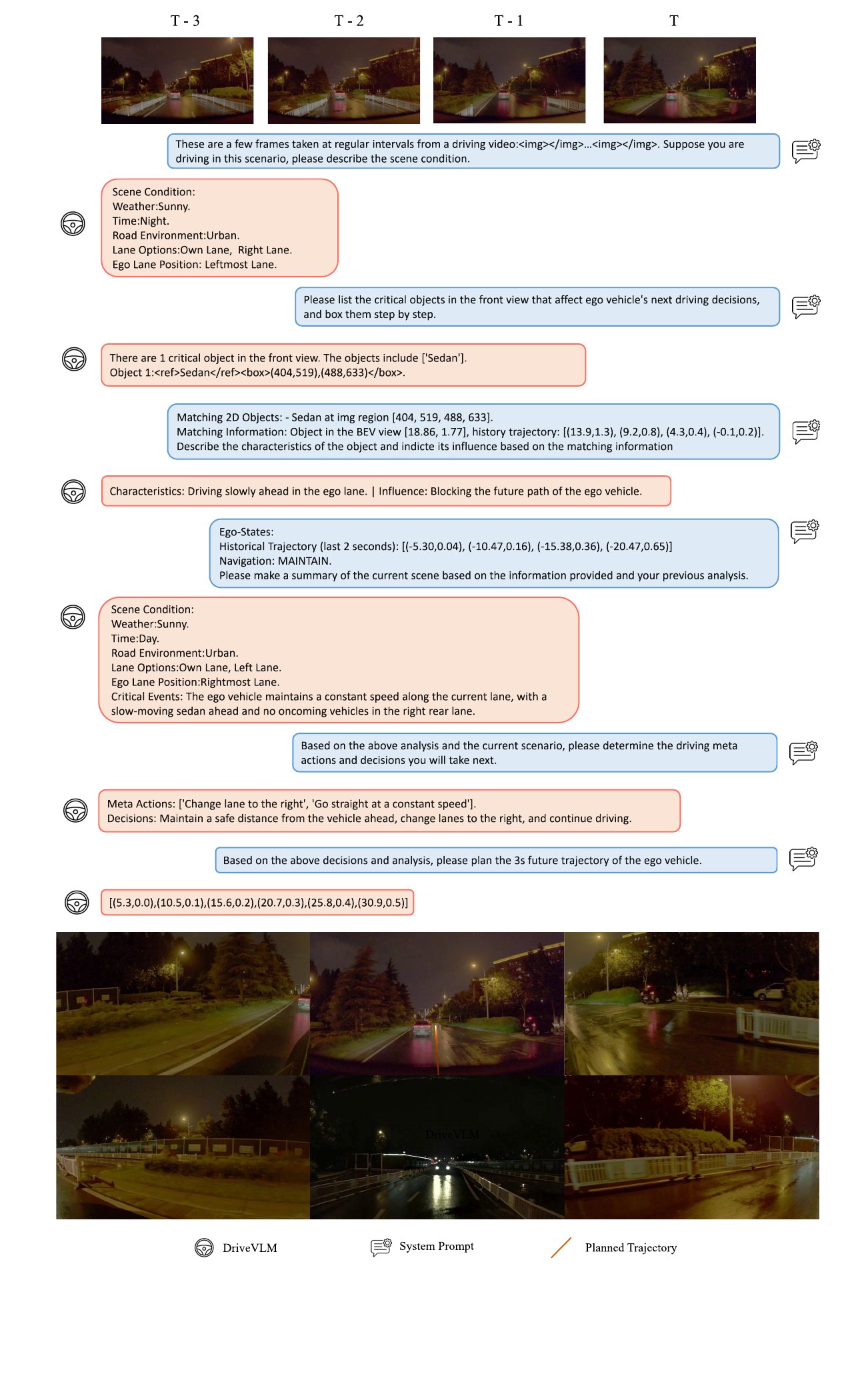}
    \vspace{-28pt}
    \caption{\textbf{Visualization of DriveVLM's output.} DriveVLM recognizes the slowly moving vehicle ahead and provides a driving decision for changing lanes to overtake.}
    \label{fig:vis-1}
\end{figure*}

\begin{figure*}[!ht]
    \centering
    \includegraphics[width=0.99\linewidth]{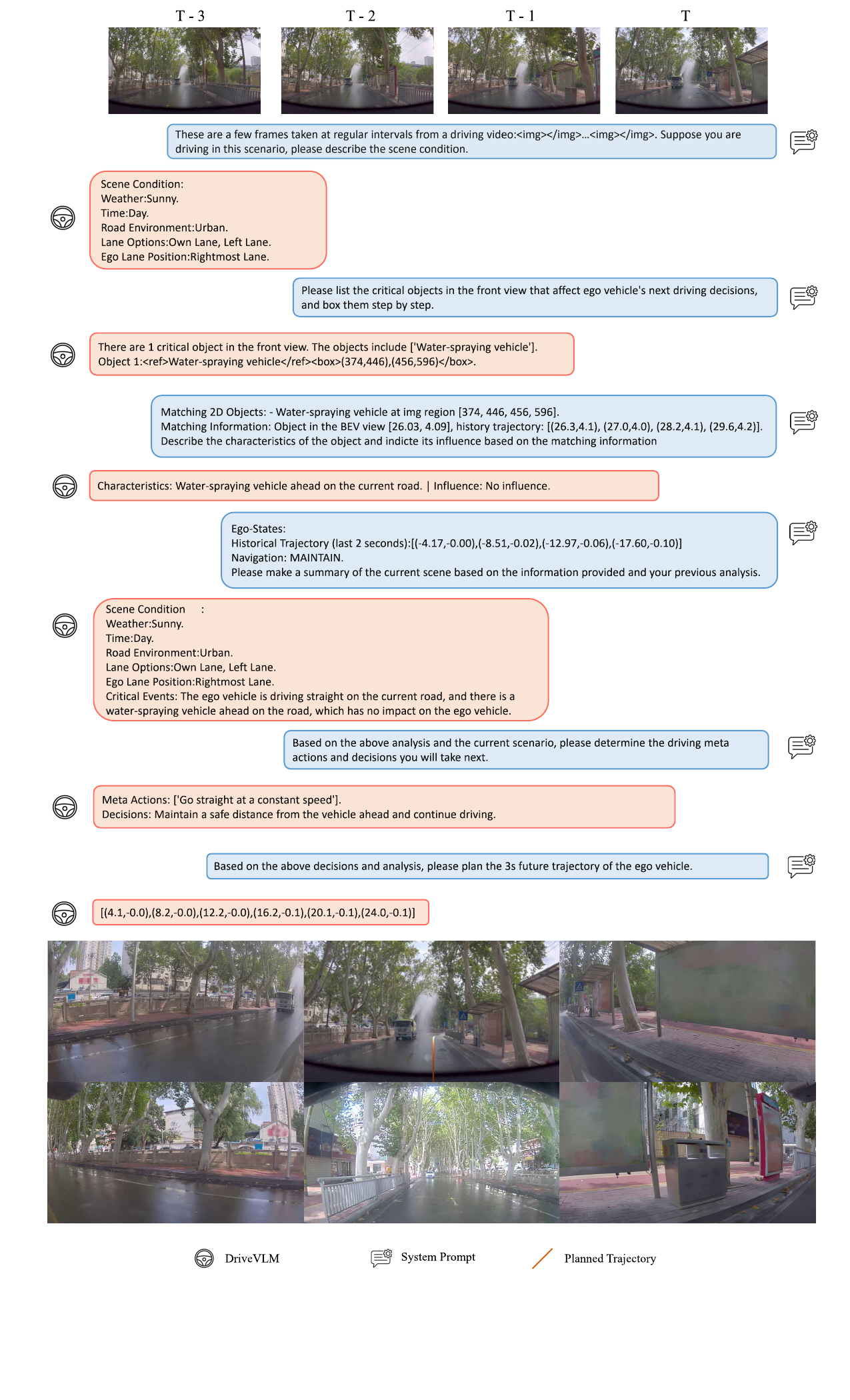}
    \vspace{-24pt}
    \caption{\textbf{Visualization of DriveVLM's output.} DriveVLM has identified the type of the unconventional vehicle ahead.}
    \label{fig:vis-2}
\end{figure*}

\begin{figure*}[!ht]
    \centering
    \includegraphics[width=0.99\linewidth]{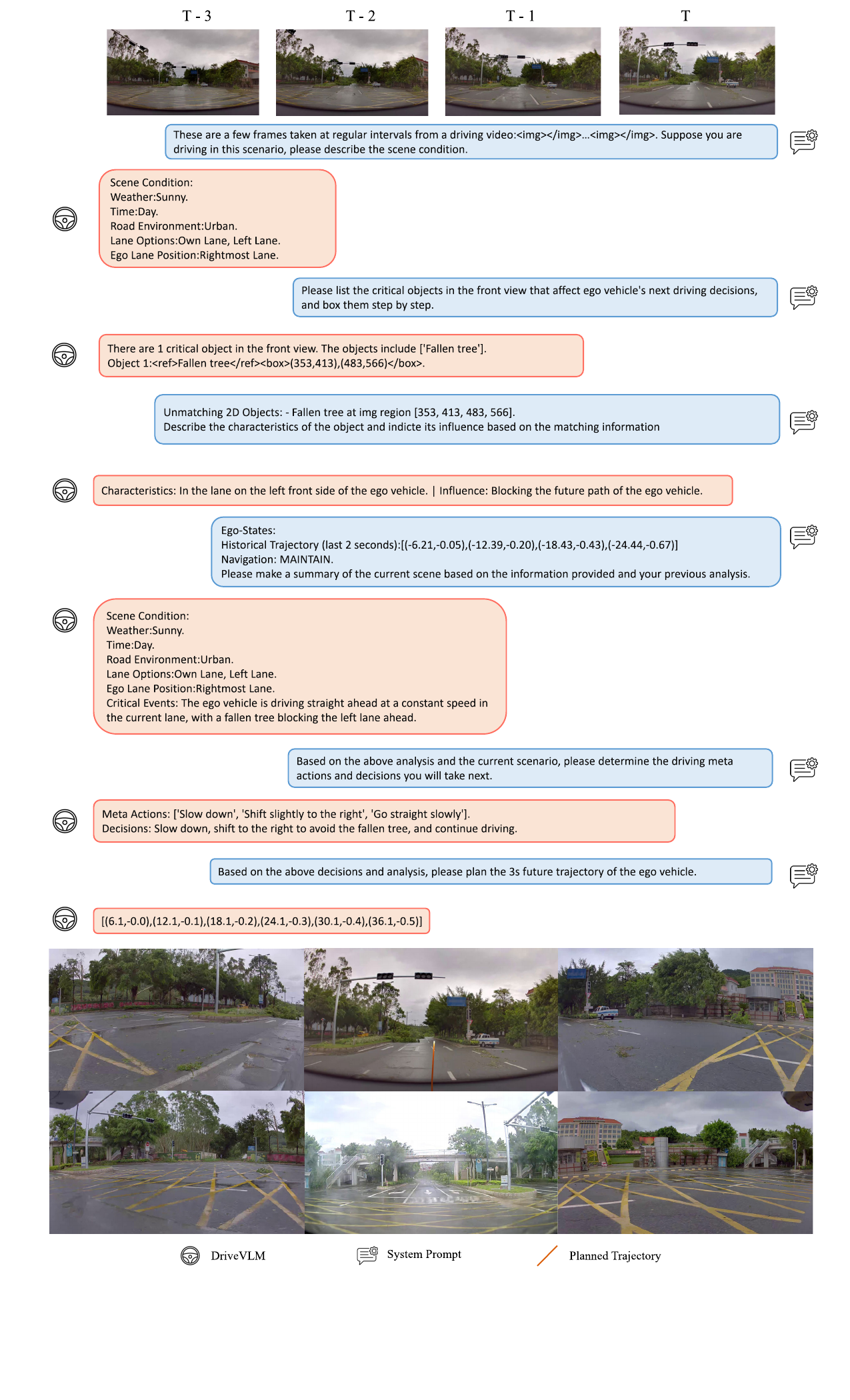}
    \vspace{-28pt}
    \caption{\textbf{Visualization of DriveVLM's output.} DriveVLM precisely detect the fallen tree and its position, subsequently planning an appropriate detour trajectory.}
    \label{fig:vis-3}
\end{figure*}

\begin{figure*}[!ht]
    \centering
    \includegraphics[width=0.95\linewidth]{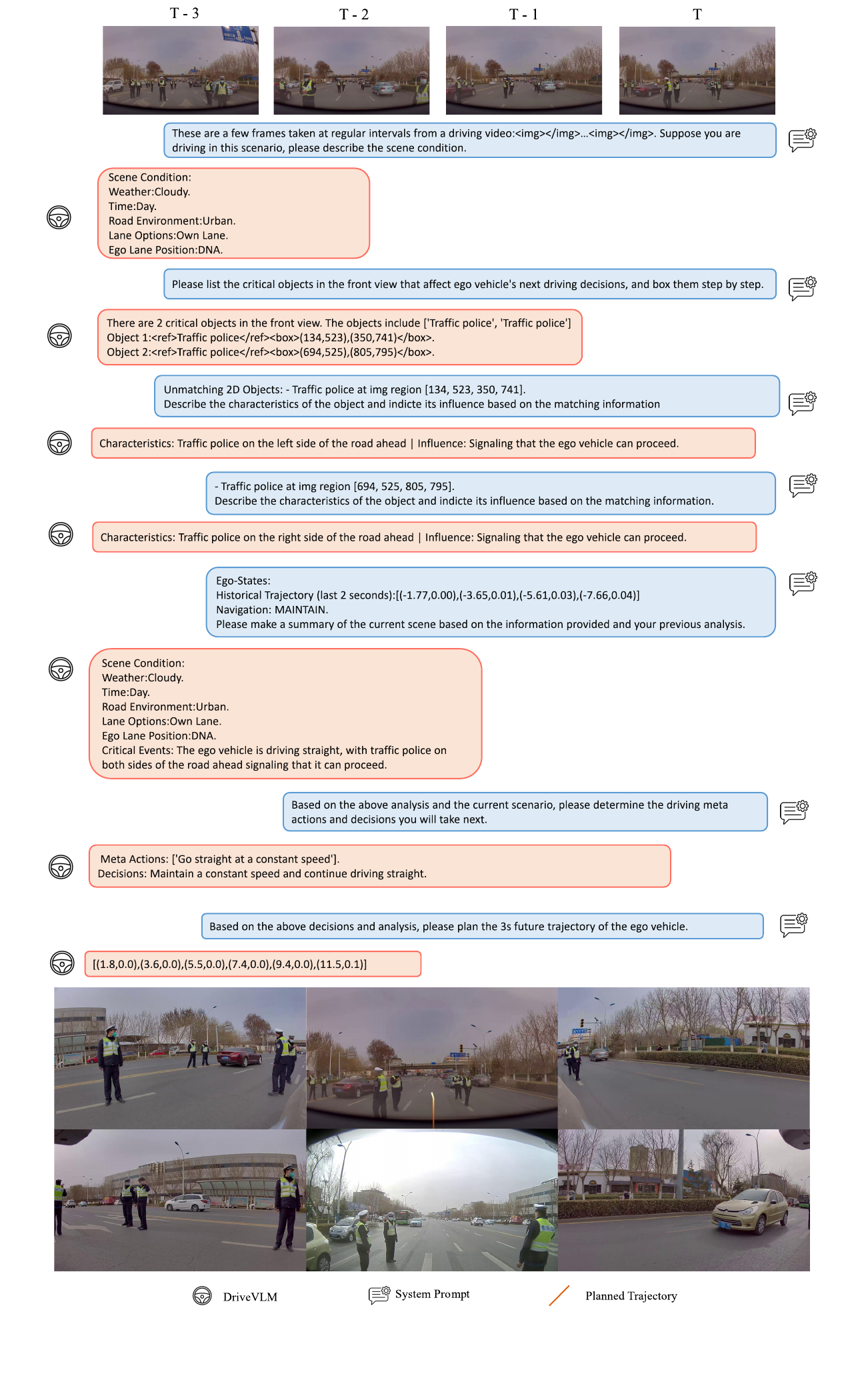}
    \vspace{-20pt}
    \caption{\textbf{Visualization of DriveVLM's output.} The traffic police signaling to proceed with hand gestures has been accurately captured by DriveVLM.}
    \label{fig:vis-4}
\end{figure*}

\begin{figure*}[!ht]
    \centering
    \includegraphics[width=0.99\linewidth]{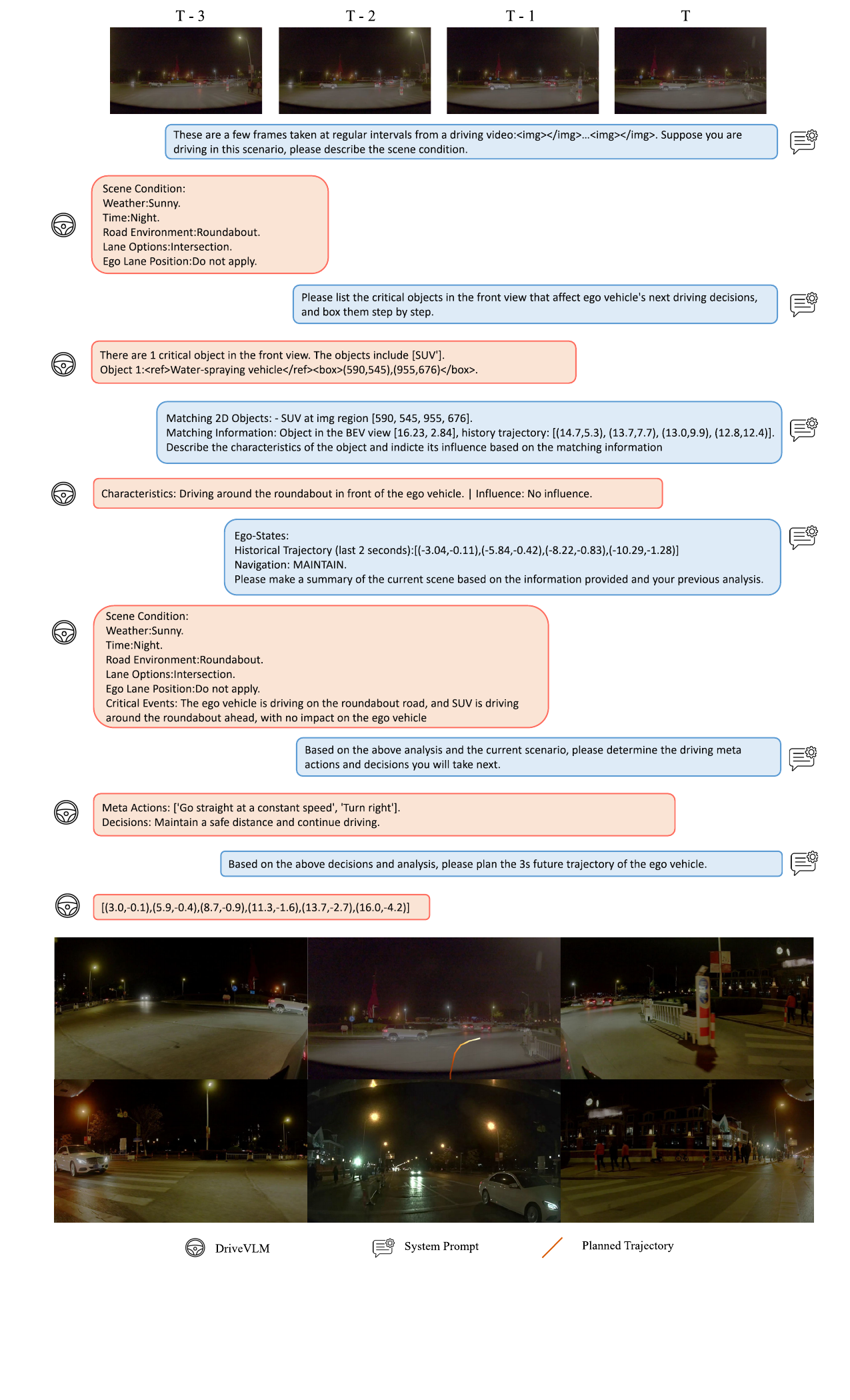}
    \vspace{-28pt}
    \caption{\textbf{Visualization of DriveVLM's output.} DriveVLM successfully recognizes the road environment of a roundabout and generates a planned trajectory with a curved path.}
    \label{fig:vis-5}
\end{figure*}

To further demonstrate the effectiveness and robustness of our DriveVLM, we provide additional visualization results in Figure~\ref{fig:vis-1},~\ref{fig:vis-2},~\ref{fig:vis-3},~\ref{fig:vis-4}, and~\ref{fig:vis-5}. In Figure~\ref{fig:vis-1}, DriveVLM recognizes the slowly moving vehicle ahead and provides a driving decision to change lanes for overtaking. In Figures~\ref{fig:vis-2}~ and~\ref{fig:vis-3}, DriveVLM accurately identifies the type of unconventional vehicles and a fallen tree, demonstrating its capability in recognizing long-tail objects. In Figure~\ref{fig:vis-4}, the traffic police signaling to proceed with hand gestures has been accurately captured by DriveVLM. In Figure~\ref{fig:vis-5},
DriveVLM successfully recognizes the road environment of a roundabout and generates a planned trajectory with a curved path.



\end{document}